\documentclass{article}

     \PassOptionsToPackage{numbers, compress}{natbib}
\usepackage[preprint]{neurips_2020}

\usepackage{natbib}
\usepackage[utf8]{inputenc} % allow utf-8 input
\usepackage[T1]{fontenc}    % use 8-bit T1 fonts
\usepackage{hyperref}       % hyperlinks
\usepackage{url}            % simple URL typesetting
\usepackage{booktabs}       % professional-quality tables
\usepackage{amsfonts}       % blackboard math symbols
\usepackage{nicefrac}       % compact symbols for 1/2, etc.
\usepackage{microtype}      % microtypography

\usepackage{amsmath}
\usepackage{amssymb}
\usepackage[dvipsnames]{xcolor}
\usepackage{graphicx}
\usepackage{multirow}
\usepackage{cancel}
\usepackage{caption,subcaption}
\usepackage[utf8]{inputenc}
\usepackage{tikz}
\usepackage{enumitem}% http://ctan.org/pkg/enumitem
\usetikzlibrary{trees}
\usepackage{color}
\usepackage[normalem]{ulem}

\newcommand{\nc}{\newcommand}
\newcommand{\mc}{\mathcal}
\nc{\beq}{\begin{equation}}
\nc{\eeq}{\end{equation}}
\nc{\bea}{\begin{eqnarray}}
\nc{\eea}{\end{eqnarray}}
\nc{\mb}{\mathbf}

\title{\textsf{dNNsolve}: an efficient NN-based PDE solver}

\author{%
V.~Guidetti\\
  DESY\\
  Hamburg, Germany\\
  \texttt{veronica.guidetti@desy.de}\\
 \And
 F.~Muia\\
 DAMTP and CTC\\
 University of Cambridge, UK\\
 \texttt{fm538@cam.ac.uk}
 \And
 Y.~Welling\\
  DESY\\
  Hamburg, Germany\\
  \texttt{yvette.welling@desy.de}\\
  \And
  A.~Westphal\\
  DESY\\
  Hamburg, Germany\\
  \texttt{alexander.westphal@desy.de}
}

\begin{document}

\begin{flushright}
DESY-21-037
\end{flushright}

\maketitle

\begin{abstract}
Neural Networks (NNs) can be used to solve Ordinary and Partial Differential Equations (ODEs and PDEs) by redefining the question as an optimization problem. The objective function to be optimized is the sum of the squares of the PDE to be solved and of the initial/boundary conditions. A feed forward NN is trained to minimise this loss function evaluated on a set of collocation points sampled from the domain where the problem is defined. A compact and smooth solution, that only depends on the weights of the trained NN, is then obtained. This approach is often referred to as ‘PINN’, from \textit{Physics Informed Neural Network}~\cite{raissi2017physics_1, raissi2017physics_2}. Despite the success of the PINN approach in solving various classes of PDEs, an implementation of this idea that is capable of solving a large class of ODEs and PDEs with good accuracy and without the need to finely tune the hyperparameters of the network, is not available yet. In this paper, we introduce a new implementation of this concept - called \textsf{dNNsolve} - that makes use of {\it d}ual {\it N}eural {\it N}etworks to {\it solve} ODEs/PDEs. These include: i) sine and sigmoidal activation functions, that provide a more efficient basis to capture both secular and periodic patterns in the solutions; ii) a newly designed architecture, that makes it easy for the the NN to approximate the solution using the basis functions mentioned above. We show that \textsf{dNNsolve} is capable of solving a broad range of ODEs/PDEs in $1$, $2$ and $3$ spacetime dimensions, without the need of hyperparameter fine-tuning.
\end{abstract}

\section{Introduction}
The modern era of the physical description of nature is built upon a very fundamental realization starting with Galilei's and Newton's work -- namely that the dynamical laws of nature are given in general by systems of coupled Ordinary and Partial Differential Equations (ODEs and PDEs), that is the laws are \emph{local} in space-time. Since then, it is also clear that describing the state of a physical system and/or predicting its time evolution into the future requires specifying appropriate boundary and/or initial conditions, and then solving the PDEs on the relevant spatial or space-time domain. 

Attempting to do so shows very quickly, that finding exact analytical solutions for any situation involving more than a few degrees of freedom of the system in question becomes prohibitively difficult in most cases. Therefore, solving PDEs for most problems became critically dependent on developing approximation schemes relying on discretizing space-time and converting the PDEs in finite-difference equations, which can then be solved \emph{algorithmically} in an iterating series of steps. This process eventually led to the development of finite-difference and finite-element method (FD/FEM) based discretized integrators.
These discrete methods of integrating ODEs and PDEs were developed into standardized software packages being able to produce numerical adaptively discretized solutions to ODEs and PDEs with controllable approximation errors for many types of ODEs and PDEs from physics and mathematics. From hereon, we will use FD/FEM to summarily denote both the PDE solvers as well as the ODE solvers. 

Finite difference methods are clearly more straightforward to be implemented. This is the main reason why they are widespread in the scientific community, see for instance~\cite{Felder_2008, Frolov_2008, Sainio:2009hm, Huang:2011gf, Sainio:2012mw, Child:2013ria, Clough:2015sqa, Mewes_2018, Giblin:2019nuv, Vogelsberger:2019ynw} for some examples of early universe, general relativity and large scale structure simulations. On the other hand finite difference methods do not work well if the domain is irregular, which is a standard situation in many engineering contexts, in which one has to deal for instance with structural analysis of buildings or deformation and stresses of solid bodies. There is a number of finite element based softwares available on the market, such as~\cite{Mathematica, MATLAB:2010, multiphysics1998introduction} and a few open source ones~\cite{maple, bastian2020dune, AlnaesBlechta2015a, mfem}. Both finite element and finite difference methods achieve outstanding results in terms of accuracy, but share a few shortcomings: first, the discretization is unavoidably a source of error. This problem can be cured in most situations by improving the precision of the solver, i.e. increasing the number of points in the grid for the finite difference methods and refining the mesh and/or choosing a basis of higher order polynomials to approximate the solution in the finite element case. Of course, this comes at the cost of larger simulation times. However, the second and probably most important problem of these methods is that they are very memory expensive, as the amount of information to be stored at any time is proportional to the number of points in the grid for finite difference methods, and proportional to the number of elements of the mesh for finite element methods. Sometimes, this comes at the expense of not being able to use faster processors, e.g. GPUs instead of CPUs, to perform large simulations. 

More fundamentally, the approximation quality of FEM integrators depends on the cleverness of the chosen higher-order discretization schemes of the solving function and its space-time derivatives entering the given ODE/PDE as well as the choice of the interpolating functions used to improve the discretized integration method. While they have become quite good, they were developed by humans based on certain criteria and motivations, which prevent these methods from being adapted from the ground up to the specific ODE/PDE in question. Instead, the methods developed for the FD/FEM paradigm have a given fixed algorithmic form but are set up to deliver reasonably accurate solutions to many ODEs/PDEs.

Thus, it is conceivable to achieve a potentially significant improvement over the FD/FEM solvers, if the functional form of the approximation algorithm itself could become dynamically adaptable to each given ODE/PDE at hand. For human programmers this would imply to re-invent a new FD/FEM algorithm each time -- hence it is here where the specific properties of machine learning based on deep neural networks (DNNs) become useful. Based on the universal approximation theorem~\cite{cybenko1989approximation, HORNIK1991251, pinkus_1999}, a DNN can act as a universal regressor which is able to represent any bounded continuous function, given suitable discretized training data, without pre-specifying a human-developed symbolic algorithm discretizing the function. 

\subsection{Relevant literature}
\label{sec:Literature}

The interplay between neural networks and differential equations presumably started back in the early '90s, when an algorithm based on the Hopfield network combined with finite difference methods was proposed in~\cite{Lee1990NeuralAF}. The idea to use a deep neural networks was then put forward in~\cite{Lagaris_1998, 870037, Lagaris:1997ap} in the second half of the '90s. The idea of these papers was very simple: a neural network can be used to approximate a function as well as its derivatives, and it can be trained in order to minimize a loss function vanishing when the PDE is solved. At that time, backpropagation was not widespread yet, hence the authors had to restrict to a quite simple architecture with a single hidden layer, in order to be able to compute derivatives analytically.

From here it is short step to the PINN, introduced in 2017 by~\cite{raissi2017physics_1,raissi2017physics_2} -- the physics-informed neural network. The idea is simple -- you use the learning and approximating power of the DNN to adjust the learned function until it satisfies a given ODE or PDE precisely enough. This is achieved by using the square of the differential equation itself, evaluated on the DNN output (its best guess of the function at a given algorithmic step called `epoch'), as the `bulk' loss function whose gradients then tell the DNN how to adjust its weights using backpropagation. The full loss function here contains separate pieces which enforce a solution of the PDE in the bulk space-time as well as fulfillment of the initial and boundary conditions determining the solution. Such a PINN will in principle learn an optimized approximate solution of the ODE/PDE given to it, instead of using a fixed symbolically prescribed `one-size-fits-all' algorithm. 

This may lead to gains in the approximation quality and/or efficiency of the dynamically learned algorithm compared to pre-fixed algorithm, and has the potential to become highly memory-conserving. A PINN, once trained to solve a given ODE/PDE only needs to store the architecture of its DNN and its trained weight system (one set of float valued matrices). On the other hand, a conventional solver has to build the discretized function system and its derived higher-order interpolating functions and, moreover, needs to maintain this information in memory during work 

An interesting alternative approach to PINNs, named the CINT (from Constrained Integration) algorithm, was proposed in 2014 in~\cite{rudd2013constrained, RUDD2015277}, based on a combination of the CPROP algorithm~\cite{ferrari2008} with Galerkin methods~\cite{galerkin1915electrical, michlin1962variationsmethoden, vajnberg1973variational}.  The CINT algorithm was shown to successfully solve the wave and heat equation, and achieves very good results both in terms of speed and accuracy. 

We now return to a closer discussion of the PINN. As already mentioned, the loss function of a PINN contains several pieces, which separately enforce the PDE solution in the bulk and its initial and boundary conditions. Using rather large networks (such as for instance a 9-layers NN containing 20 units for each layers for the Burgers' equation, for a total of more than 3000 parameters), the authors of~\cite{raissi2017physics_1} achieved quite good results, reaching a loss $< 10^{-3}$ for a variety of PDEs, such as the Burgers'  and the Schr\"odinger equations in $1+1$D. 

A more sophisticated architecture was proposed under the name of \textit{Deep Galerkin Method} (DGM) in 2018 in~\cite{Sirignano_2018} with the aim of solving high-dimensional PDEs. The architecture is reminiscent of LSTM models, and is designed to successfully detect sharp turns and in the solution. The DGM architecture makes use of distributed training and a Monte-Carlo method for the computation of second derivatives to solve various high-dimensional (up to 200 dimensions) PDEs such as the Burgers' equation and the Hamilton-Jacobi-Bellman equation.

More recently, various packages exploiting these ideas have been put forward. For instance, \textit{DeepXDE} introduces a residual-based adaptive refinement of the grid in order to improve the training efficiency of PINNs~\cite{lu2020deepxde}. DeepXDE is shown to successfully solve PDEs that are first order in time derivative as well as integro-differential equations and inverse problems. Very nicely, the DeepXDE Python library can be easily adapted to work on non-trivial domain shapes. On the other hand, \textit{Pydens}~\cite{koryagin2019pydens} is a Python library that allows to solve a large family of PDEs including the wave equation and the heat equation, and gives the possibility to easily switch between standard architectures (such as ResNet and DenseNet).

Beyond the ideas mentioned above, there are many more that would need to be mentioned, including: the use of adaptive activation functions to accelerate the convergence of PINNs~\cite{jagtap2020adaptive, jagtap2020locally}, the use of convolutional neural networks to solve PDEs or to address the inverse problem~\cite{tompson2017accelerating, Geneva_2020, Thuerey_2020}, adversarial training techniques~\cite{Yang_2019}, energy based approaches~\cite{Samaniego_2020}, the use of random approximation methods~\cite{E_2017, chaudhari2017deep, Beck_2019, fujii2019asymptotic}. Note that it has been estimated that techniques based on PINNs might become competitive with standard methods such as finite difference~\cite{avrutskiy2020neural}. See also~\cite{dockhorn2019discussion} for a recent discussion of the advantages and disadvantages of using neural network based techniques for solving PDEs and about the possible routes forward in this direction.

\subsection{Our contribution}

At the moment though, we are not able to find a single implementation of the PINN or any of the other ideas that is capable of solving a large class of ODEs and PDEs with good accuracy and without the need to finely tune the hyperparameters of the network. By `large class of ODEs and PDEs’ we mean to include not only the simplest, mandatory examples, such as the wave equation, the heat equation and the Poisson equation, but also more complex examples that include for instance damped oscillations and stiff equations. 

The crucial idea we report here, which makes the PINN more efficient to use for solving many different ODEs/PDEs, is to develop a DNN architecture which optimizes the set of functions used in the DNN's exploitation of the universal function theorem. Namely, we first observe that bounded solutions of any ODE/PDE can in principle be decomposed into any given complete orthonormal function system. Fourier decomposition and the softened-bin decomposition of a function, used to prove the universal approximation theorem for feed-forward DNNs, are two examples. 

In our experience in physics, the various types of dynamics shown by ODE and PDE solutions, the resulting function very often is composed out of a secular non-oscillatory part, and a part which is a Fourier synthesis of various oscillatory components. While it is possible to approximate a secular slow-changing function solely by Fourier synthesis, doing so requires summing over a very large number of Fourier modes to achieve satisfactory accuracy. Conversely, approximating an oscillatory function over many periods using the softened-bin decomposition requires summing up a very large number of narrow bins. Both decompositions are possible, although inefficient, for the each of two the respective cases. 

Hence, for achieving a given approximation quality, we expect it to be much more efficient, to use both the secular behaviour of the function system used to build the softened-bin decomposition and Fourier decomposition -- the first to efficiently approximate the secular components of a function, the second to model its oscillatory components. Such a `two-stream' decomposition will likely require far fewer terms/modes to approximate a function to given finite accuracy, than each of the two decomposition streams by itself. For a similar approach in the context of time-series analises, see for instance~\cite{TimeSeries}.

Therefore, we propose \textsf{dNNsolve}: the `two-stream' decomposition into the DNN of the PINN by splitting some of the hidden layers into two parts whose choice of activation function is suitable for the non-oscillatory secular decomposition in one part of the layer, and for a Fourier decomposition in the second part of the layer (using periodic activation functions that were previously used for instance in the context of Fourier NNs~\cite{Silvescu1999FourierNN, sitzmann2020implicit}). After passing this `two-stream' layer, the two decomposition streams are algebraically combined to produce the DNN output.

We test \textsf{dNNsolve} on a broad range of ODEs and PDEs in 1, 2 and 3 dimensions. The main improvements on the one-stream PINNs are:
\begin{itemize}
\item We use the \textit{same} hyperparameters (such as learning rate, network size and number of points in the domain) for each $d$-dimensional PDE and achieve good accuracies for all of them.
\item A small NN architecture with $d\times \mathcal{O}(100-200)$ trainable parameters is effective for the problems at hand, while keeping the number of sampled points at $\mathcal{O}(1000)$. On the other hand, in e.g.~\cite{lu2020deepxde} PINNs with $ \mathcal{O}(1000-8000)$ parameters were employed for 1- and 2-dimensional examples.
\item We use only $\mathcal{O}(1000)$ epochs of ADAM (or $\mathcal{O}(100)$ with mini-batches) followed by a variable number of iterations of BFGS until convergence. This may be contrasted with the $15000-80000$ epochs of ADAM used in~\cite{lu2020deepxde}.
\end{itemize}

The plan of this note is as follows. In section~\ref{sec:Architecture} we will proceed to describe the setup of the two-stream PINN, and discuss its loss function as well its weight initialization and hyperparameter choices.
In section~\ref{sec:Results}, we present its application to significant number of ODEs and PDEs and (0+1), (1+1) and (2+1) dimensions. To gauge the accuracy of the solutions learned, we compare an accuracy estimate based on the root mean squared (RMS) error provided by the PINN loss function as well as the RMS error between the PINN solution and the known analytical solution for each example. A lot of the technical details are relegated to several appendices~\ref{sec:AppendixA}, \ref{app:1D}, \ref{app:2D}, \ref{app:3D}.

\section{Dual base network}
\label{sec:Architecture}

The aim of our work is to find an improved PINN architecture that is able to solve a general differential equation~\cite{Lagaris:1997ap}
\beq
G(\vec{x},u^{l}(\vec{x}),\nabla u^{l}(\vec{x}), u^{l}(\vec{x}),\nabla^2 u^{l}(\vec{x}), u^{l}(\vec{x}),\dots)=0\;,\qquad \vec{x}\in \mathcal{D} \subset \Bbb R^{D}\;, \qquad  \vec{u}(\vec{x})\in \Bbb R^{n_{\rm o}} \,,
\label{eq:PDE}
\eeq
subject to certain boundary and initial conditions, where $\mathcal{D}$ is the domain, $D$ is the number of spacetime dimensions and $n_{\rm o}$ is the number of components of $\vec{u}$. Therefore we look for a scalable architecture that can be easily adapted to problems having different input and output domain dimensions. In order to do this, we take direct inspiration from series expansion. The universal approximation theorem in its ‘arbitrary width’ formulation~\cite{cybenko1989approximation, HORNIK1991251, pinkus_1999} states that a feed forward NN with a single hidden layer and arbitrary continuous activation functions can approximate every continuous functions with arbitrary precision. On the other hand, carefully choosing the architecture and activation functions may be useful for computational purposes, reducing the amount of neurons needed to get the solution in a finite amount of time and with finite computational power.

\begin{figure}[h!]
\begin{center}
  \includegraphics[scale=0.67]{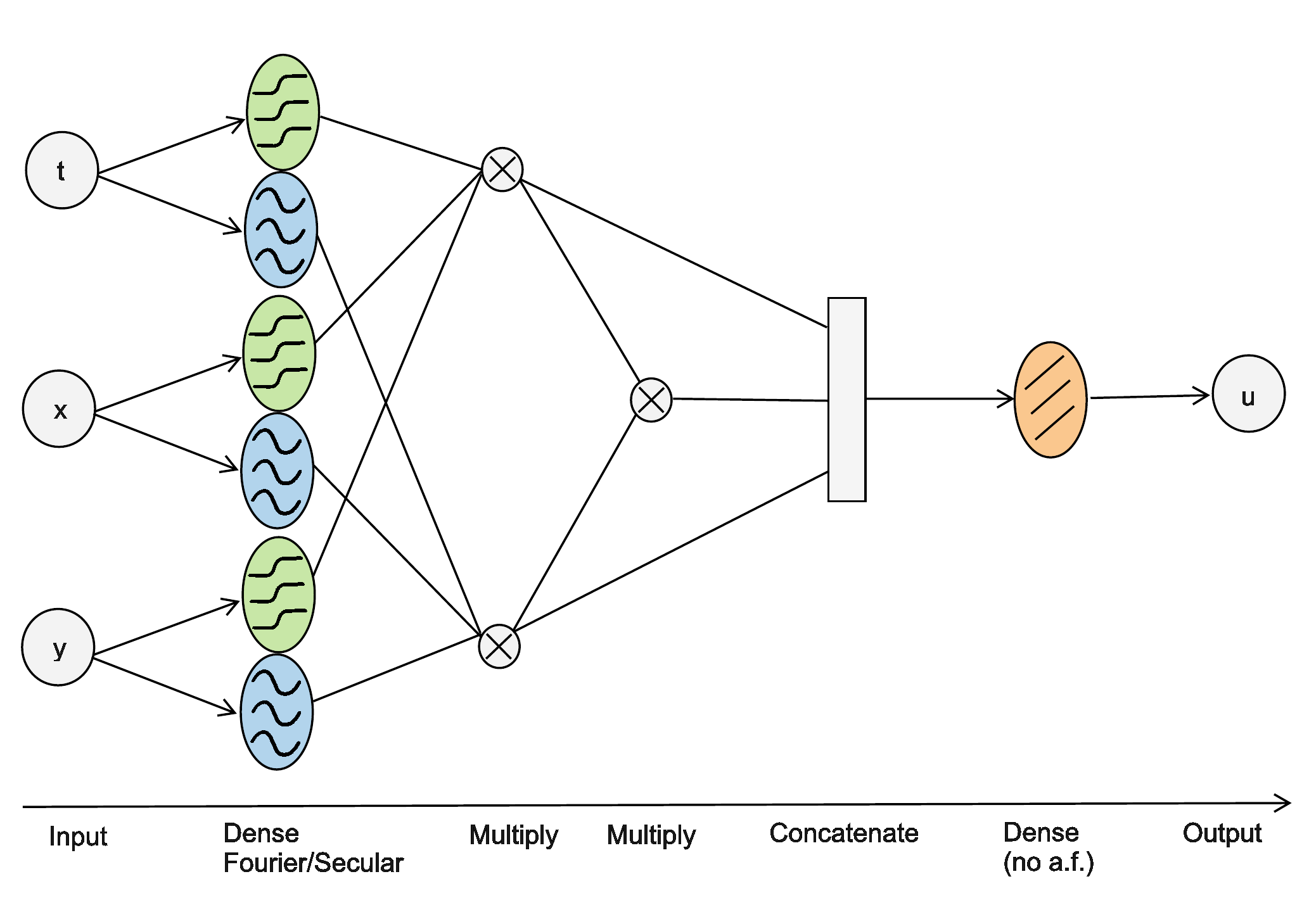}
  \caption{\textsf{dNNsolve} network architecture for the case $D = 3$: multiple input variables are processed independently using Dense layers with sine and sigmoid activation functions. The results are then processed by three multiplication layers and finally merged together using a concatenation layer. The final output (scalar in this picture) is then computed using a simple dense layer with no activation function.}
  \label{fig:architecture}
  \end{center}
\end{figure}

One of the best-known tools to perform function decomposition is given by the Fourier series that allows one to decompose each piece-wise continuous function on some given interval. Nevertheless, also in this case, an exact function decomposition may be given by an infinite sum of  harmonic terms (a simple example is given by the infinite series representing a straight line). Trying to reduce the number of nodes required to obtain a satisfactory approximation of our PDE solution, given that we can not know a priori its behaviour, we decide to use two kinds of activation functions, i.e. sine and sigmoid functions. The latter can be used to reduce the number of neurons used in capturing the secular variations of the solution, while the sine performs a Fourier-like decomposition. We combine the two types of activation functions both linearly and non-linearly, trying to improve the robustness and flexibility of our network.
Our network architecture, dubbed \textsf{dNNsolve} is given by a feed forward NN with multiple inputs and outputs, five hidden layers (of which only two layers contain trainable parameters) and two branches representing the periodic and non-periodic expansion of the solution. In the first layer each input variable is processed independently by two dense layers having the same number of neurons $N$ with sigmoidal and sinusoidal activation function, respectively:
\begin{center}
\begin{tikzpicture}[level distance=1.5cm,
  level 1/.style={sibling distance=6cm},
  level 2/.style={sibling distance=1.5cm}]
  \node {$x^i$}
    child {node {$  \sin(\omega_k^{(i)} x^i  + \phi_k)\equiv f_k^{(i)}$}}
    child {node {$  \sigma(w_k^{(i)} x^i  + b_k)\equiv s_k^{(i)}$}}
    ;
\end{tikzpicture}
\end{center}
where $k=1, \dots, N$ and $i = 1, \dots, D$, where $D$ is the number of input variables, i.e. the number of spacetime dimensions of the problem.
Afterwards, the periodic and non-periodic outputs are merged together using three multiplication layers, see Figure~\ref{fig:architecture}. These perform the element-wise multiplication of the neurons belonging to different input variables. This operation is performed for the two branches separately and for their product to take into account some non-linear interplay between the periodic and non-periodic decomposition of the solution.
The output of the multiplication layers, identified by $\mb{F}$, $\mb{S}$ and $\mb{FS}$, can be summarized as:
\beq
\label{eq:HiddenLayers}
\mb{F}=\{F_k\}=\bigg\{\prod_{j=1}^{D}f_k^{(j)} \bigg\}\;, \qquad \mb{S}=\{S_k\}=\bigg\{\prod_{j=1}^{D}s_k^{(j)} \bigg\}\;, \qquad \mb{FS}=\{F_k S_k\} \,,
\eeq
where $k=1,\dots, N$.
These results are then joined together using a concatenation layer that is taken as input by a dense layer, $\mb{D}$, with no activation function having $n_{\rm o}$ neurons, where $n_{\rm o}$ is the number of unknown functions we need to compute as output of our network (this depends on the problem at hand): $\mb{D}_l(y)=d_{il} y^i + a_{l}$,  $l=1, \dots, n_{\rm o}$.
In the case of a single ODE with scalar input $x$, the output of our network is a single function $\hat{u}$. In this case, $D = 1$ and $n_{\rm o} =1$, and the output can be written as:
\beq
\label{eq:ExplicitODE}
\hat{u}(x)=\sum_{k=1}^{N} d_k \sin(\omega_k x  + \phi_k) +  d_{N+k}\sigma(w_k x + b_k) + d_{2N+k}\sin(\omega_k x  + \phi_k)\sigma(w_k x + b_k) + a\,,
\eeq
where we have denoted $d_{k1} \equiv d_k$ and $a_1 \equiv a$.
A pictorial representation of our architecture for multiple inputs can be found in Figure~\ref{fig:architecture}.

Let us also mention that we tried several alternatives to the sine and sigmoid activation functions. In the secular branch of the network we also tried for instance with several possible activation functions: tanh, Gaussian, Hermite polynomials, $\tanh(x^p)$. In the Fourier part of the network we made a few experiments with wavelet layers~\cite{zhang1995wavelet}. Overall, we observed that the combination of sinusoidal activation functions with sigmoids is the best choice both in terms of accuracy and in terms ot training efficiency.

\paragraph{Weight initialization}
The number of network parameters is given by $4ND + n_{\rm o} (3N + 1)$, it scales linearly both in the number of input variables $d$ and in the number of neurons per branch $N$. After trying different possibilities, we choose the following weight initialization procedure: $d_k=10^{-4}$, $w_k^{(i)}\sim \mc{U}(0,10^{-3})$, $\omega_k^{(i)}\sim \mc{U}(\pi/(x_i^{\rm max}-x_i^{\rm min}),N\pi/(x_i^{\rm max}-x_i^{\rm min}))$ where $N$ is the number of neurons per branch. Biases, $a_l$, $\phi_k$, $b_k$, are always initialized to zero. The choice of the initialization for the frequency of the sine activation functions reflects some prior knowledge about the physical scales that characterize the system: for instance in the one-dimensional case, we do not expect hundreds of oscillations in the domain of definition of the ODE. The lowest frequency is chosen to be first harmonic since we want the sine activation functions to capture only the oscillating part of the solution as lower frequencies can be easily captured by sigmoid neurons. Regarding the amplitude parameters $d_i$, we choose to initialize them to tiny values, i.e. we start from nearly vanishing solution. This decision is related to the behaviour of the bulk contribution to the loss function. In case the ODE/PDE is homogeneous and admits a zero solution, in the first steps of training the algorithm tends to drive $d_i$ to zero before it starts learning the real solution. This happens because $\hat{u}=0$ appears to be a local minimum of the loss function. The amplitude initialization we choose is then able to shorten training under certain conditions that we describe in the next section.
Our choice of processing input variables independently, instead of using them as a single input for a dense layer, is motivated by empirical observation. Dense layers with sinusoidal activation function over linear combinations of inputs seemed to be less flexible during training, i.e. the system got systematically stuck in local minima. In order to overcome this problem, fine-tuned frequency weights initialization was required, leading to a poor level of generalization of the network. As opposed to sigmoid activation functions, sinusoidal functions in 2/3D are non localised in space and do not adapt well: a small change in one parameter implies changing the function in the whole area/volume. This makes their training very hard since the loss function turns out to be fulfilled with local minima~\cite{819741, Parascandolo2016}. We empirically found that this behaviour can be mitigated by processing input variables independently. This may be due to the presence of extra phase parameters ($\phi_k$) or because we reduce the effects of changing weights to a single dimension. 
\paragraph{Loss function}
Following~\cite{Sirignano_2018} we use a piece-wise loss function, given by the sum of different contributions coming from bulk, boundary and initial conditions points. Let us write the domain of our problem splitting it into time and space components: $\mathcal{D}=[t_0,t_L]\times \Omega$ where $\Omega\subset \Bbb R^{D-1}$. Different PDEs, see Eq.~\eqref{eq:PDE}, having different degrees in time and space variables can be subject to different conditions. In the examples showed in this work we mainly considered

\begin{itemize}
\item Initial conditions:
\begin{equation}
u^l(t_0,\vec{x})=u_0^l(\vec{x}) \,, \qquad [\partial_t u^l](t_0,\vec{x})=v_0^l(\vec{x})
\end{equation}
\item Boundary conditions:
\begin{eqnarray}
&& u^l(t,\vec{x})|_{\partial\Omega}=f^l(\vec{x}) \qquad \quad  \text{(Dirichlet)} \,, \\
&& \left[\partial_{\vec{n}} u^l (t,\vec{x})\right]|_{\partial\Omega}=g^l(\vec{x}) \quad \text{(Neumann)} \,,
\end{eqnarray}
where $\vec{n}$ is the normal to the boundary unit vector.
\end{itemize}

Collectively denoting by $\theta=\{w,\omega,\phi,b,d,a\}$ the set of NN parameters and $\hat{u}^l(\vec{x},t,\theta)$ the output of the NN, the general shape of the loss function is given by:
\beq
\label{eq:Loss}
\begin{array}{lll}
\mc{L}(t,\vec{x},\theta,\alpha) &\equiv \mathcal{L}_{\Omega} + \mathcal{L}_{0} + \mathcal{L}_{\partial\Omega} = \\
&\displaystyle =
\sqrt{
\frac{1}{n_\Omega}\sum_i \left[
G(t_i,\vec{x}_i,\hat{u}^{l}(t_i,\vec{x_i},\theta),\nabla \hat{u}^{l}(t_i,\vec{x_i},\theta),\dots)\right]^2} \,+ \\
&\displaystyle + \, \alpha_0 \sqrt{
\frac{1}{2 n_0}\sum_j \left(\left[\hat{u}^l(t_0,\vec{x}_j,\theta)-u_0^l(\vec{x}_j)\right]^2+ \left[[\partial_t \hat{u}^l](t_0,\vec{x}_j,\theta)-v_0^l(\vec{x}_j)\right]^2\right)} \, +\\
&\displaystyle + \, \alpha_{\partial\Omega} \sqrt{\frac{1}{n_{\partial\Omega}} \sum_k \left[u^l(t_k,\vec{x}_k,\theta)|_{\partial\Omega}-f^l(t_k,\vec{x}_k) \right]^2} \,,
\end{array}
\eeq
where we considered Dirichlet boundary conditions and second order PDEs in time. We take the sum of square roots instead of the sum of squares, because empirically we observe 
that this typically converges faster and is more accurate.
If the boundary condition is not Dirichlet, the last term of the loss should be modified accordingly. Also, if the PDE is first order in time, the second term in the root of the second line of Eq.~\eqref{eq:Loss} disappears. The loss function is computed over a batch of size $n_s=n_\Omega+n_{\partial\Omega}+n_0$ where $n_\Omega$, $n_{\partial\Omega}$ and $n_0$ are the number of points belonging to the bulk, the boundary and the initial condition respectively. Our loss function depends on two parameters, $\alpha_0 $ and $\alpha_{\partial\Omega}$ that can be considered as additional hyperparameters of the neural network. This kind of approach was also used in other works as~\cite{lu2020deepxde, Shin_2020}. These parameters are needed to let the NN be able to get out of local minima. For example, consider the case where an homogeneous PDE contains an identically zero solution and boundary conditions set the function to be vanishing on $\partial \Omega$. If we consider $\alpha_0=\alpha_{\partial\Omega}=1$ we have that 2/3 of the loss function drive the solution toward the local minimum $u\equiv 0$. From empirical observations we saw that the extent of this problem increases with the number of input dimensions $d$. On the other hand, taking $\alpha_0\gg 1\,,\,\alpha_{\partial\Omega}$ will introduce a hierarchy in $\mc{L}$, pushing the solution away from zero (see~\cite{vandermeer2020optimally} for a work in an analogous direction). These weights can in principle also be used to put some priority in the piece-wise loss function: if all weights are equal and different parts of the loss function do not share local minima or saddle points, the convergence rate is slowed down. Some intuitions about the relation between different pieces of the loss function and the role played by $\alpha$ hyperparameters can be found in App.~\ref{sec:AppendixA}.
\paragraph{Optimizer and training}
Loss functions often exhibit pathological curvature, saddle points and local minima. Some of these problems can be addressed using adaptive SGD algorithms, as ADAM.  The use of mini-batches represents a source of stochasticity that helps in getting out of local minima, while momentum is able to get away from saddle points and proceed in pathological curvature regions accumulating the gradient over the past steps to determine the direction to go. Moreover, momentum gives a one-shot chance of getting out of local minima. As explained in the previous paragraph, the problem we aim to solve may suffer from the presence of local minima. Some of these can be due to discretization, others derive from the definition of the loss function, see App.~\ref{sec:AppendixA}. For this reason we decide to perform a two-step training using different optimizers.
In the first step we use the ADAM~\cite{kingma2014adam} optimizer for a fixed amount of epochs using mini-batches of 256 data. We start from a huge learning rate $\lambda=0.1$ and use the standard Tensorflow learning rate decay callback with parameter choices which reduce $\lambda$ by a factor $\delta=0.5$ when the loss does not decrease by an amount of $10^{-4}$ in $30$ iterations. ADAM training should end in a neighborhood of the global minimum. In the second step, we use BFGS algorithm~\cite{Dennis:1974:CSC,GoodBengCour16}, a second order quasi-Newton method that estimates the curvature of the parameter space via an approximation of the Hessian. This is able to speed up the convergence rate, reaching the minimum in a small number of iterations. We use BFGS without using mini-batches so as to have the best approximation of the loss volume. Training ends when the supremum norm of the gradient vector is below $10^{-8}$.

\paragraph{Validation of the results} In order to be able to check the accuracy of the results, we mainly considered PDEs with analytic solutions, with a few notable exceptions that will be discussed in due course. The accuracy of the solution will be expressed in terms of the root mean squared error $r$, that in the case of an initial/boundary value problem is defined as
\beq
r = \displaystyle \sqrt{\frac{1}{n_{\rm tot}} \sum_{i} \left|\hat{u}(t_i, \vec{x}_i, \tilde\theta) - \tilde{u}(t_i, \vec{x}_i)\right|^2} \,,
\eeq
where $\hat{u}(t, \vec{x}, \tilde{\theta})$ is the NN output evaluated using the trained weights $\tilde\theta$, while $\tilde{u}(t, \vec{x})$ is the analytic solution evaluated at the point $(t, \vec{x})$. The $n_{\rm tot}$ collocation points $(t_i, \vec{x}_i)$ are taken on a regular grid with $\{200, 50, 30\}$ points per spacetime dimension in 1D, 2D and 3D respectively. Note that, even for a second order PDE in time, we do not have a separate initial condition and boundary piece, as with $r$ we only want to measure the average distance from the true solution. In the subsequent sections, we will report the loss values (both the total loss as well as the various components of the loss in Eq.~\eqref{eq:Loss}), to illustrate how the training proceeds, and the root mean squared error $r$, to evaluate the accuracy of the solutions otained using \textsf{dNNsolve}.

\section{Results}
\label{sec:Results}
We study the performance and the robustness of our architecture in solving 1D ODEs and 2D/3D PDEs. In order to do so, we use the same weight initialization in all examples. Moreover, we consider the same number of points and architecture in all equations having same dimensionality, i.e. $D$ and $n_{\rm o}$. The same criterion is applied to hyper-parameters: we try to keep them fixed within each group of ODEs/PDEs. Numerical results show that this was possible for 1D/2D problems, while 3D problems required to slightly adjust the $\alpha$ parameters for the reasons described in the previous paragraph.

Please, note that the results that we are listing in this section do not correspond to our best results, as they are obtained using a random set of initialized weights as opposed to the best solution selected out of multiple runs~\cite{lu2020deepxde}.

All the computations were performed with a laptop whose specifics are:
\begin{itemize}
\item Machine: MacBook Pro (Retina, 15-inch, Mid 2015),
\item Processor: 2.8 GHz Intel Core i7 (quad-core, 64bit, hyperthreading = 8 logical cores),
\item Memory: 16 GB 1600 MHz DDR3,
\item OS: MacOS Mojave 10.14.6.
\end{itemize}

We used \textsf{Tensorflow 2.4}~\cite{abadi2016tensorflow, dillon2017tensorflow}. All the computations were automatically parallelized via hyperthreading over 8 CPU cores. The RAM usage was always less than 500 MB for all the equations in any number of spacetime dimensions.

\subsection{1D ODEs}
\label{sec:1DODEs}

In 1D, we considered several initial value problems on the domain $t \in [0, 20]$ of the form
\begin{eqnarray}
G(t, u(t), u'(t)) = 0 \,, \qquad u(t_0) = u_0 \,, \qquad u'(t_0) = v_0 \,,
\end{eqnarray}
where the constraint on initial first derivative appears only in second order problems. In addition we solve a representative example of boundary value problem of the form
\begin{eqnarray}
G(t, u(t), u'(t)) = 0 \,, \qquad u(t_0) = u_0 \,, \qquad u(t_L) = u_L \,.
\end{eqnarray}
as well as a first order delay equation of the form
\begin{eqnarray}
G(t,t', u(t),u(t'), u'(t)) = 0 \,, \qquad u(t|t<t_0) = f(t) \,.
\end{eqnarray}
The explicit ODEs under study are listed in App.~\ref{app:1D}.
In all the examples we used $\{ n_{\Omega},n_{0}\}=\{ 2000,1\}$ collocation points and $n = 35$ neurons per branch. The oscillon profile equation, Eq.~\ref{eq:Oscillon} is a particular case: it is a boundary value problem in 1D and it is typically solved using the \textit{shooting method}. As it is particularly sensitive to the noise of unused neurons, we use only $n = 10$ neurons per branch in that particular example. It was possible to successfully solve all the equations reported in App.~\ref{app:1D} by fixing $\alpha_0=1$. The number of epochs using ADAM optimizer is $150$, while the total number of epochs, including the ones that use BFGS is reported in Tab.~\ref{tab:1D}.\\

\begin{figure}[t!]
\begin{center}
 \includegraphics[width=0.45 \textwidth]{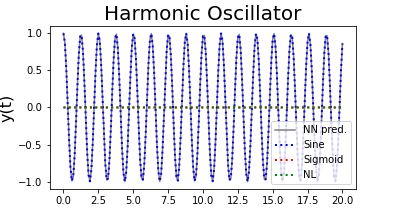}\,
  \includegraphics[width=0.45 \textwidth]{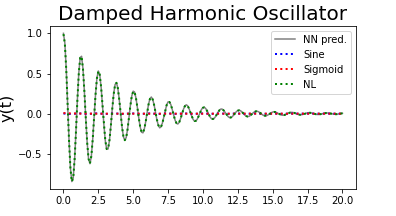}\\ \includegraphics[width=0.45 \textwidth]{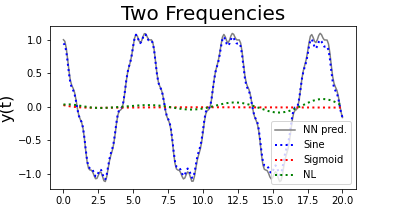}\, \includegraphics[width=0.45 \textwidth]{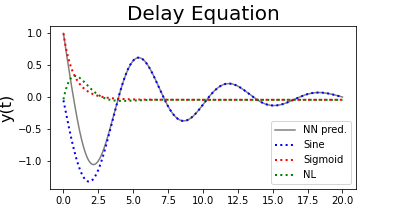}\\
   \includegraphics[width=0.45 \textwidth]{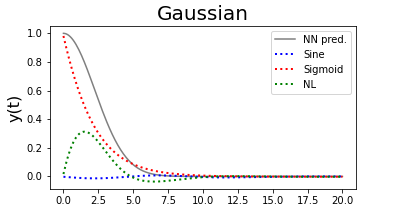}\, \includegraphics[width=0.45 \textwidth]{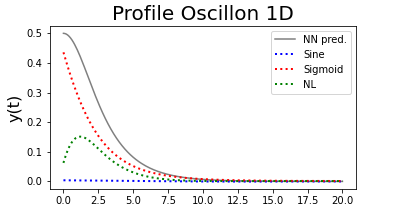}\\
  \caption{Contributions coming from sine, sigmoid and non-linear branches to the solution of six of the 1D equations considered. As expected, the harmonic oscillator and its damped version are completely solved by the sinusoidal and non-linear branches respectively. In the two frequency example the sinusoidal branch is mainly used, but as it is clear from the solution at $t > 10$, noisy contributions from unuseful neurons are not completely switched off, which calls for a better optimisation. For the Delay Equation the sinusoidal branch captures the oscillatory part of the solution, while the other branches take care of the linear part. For the Gaussian and Oscillon profile equations, the non-linear and secular branches contribute to the final solution, while the sinusoidal is switched off.}
  \label{fig:branches}
  \end{center}
\end{figure}

From our results we can see that the \textsf{dNNsolve} is able to learn different kinds of solutions without changing the setting of weight initialization and hyperparameters. The network is able to learn which function basis can better approximate the solution and reduce the contributions coming from the other branches. On the other hand, the capability of really switching off redundant degrees of freedom still needs to be optimized. Six examples showing the interplay between different branches can be found in Figure \ref{fig:branches}. We see that, despite the network is able to focus on the right basis, it happens that spurious contributions may appear from the other branches as in the ‘two frequencies equation’, Eq.~ \ref{eq:DoubleFreq}. This problem becomes more apparent when the number of neurons per branch increases.

\begin{figure}[h!]
\begin{center}
  \includegraphics[width=0.49 \textwidth]{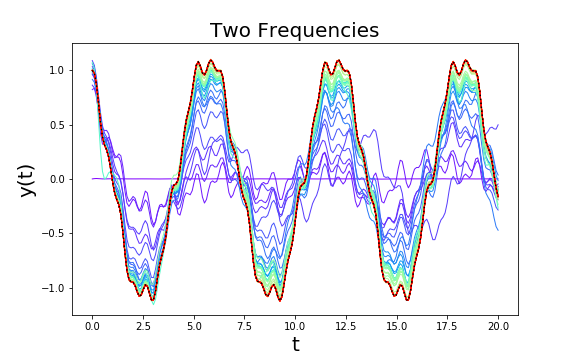}\, \includegraphics[width=0.49 \textwidth]{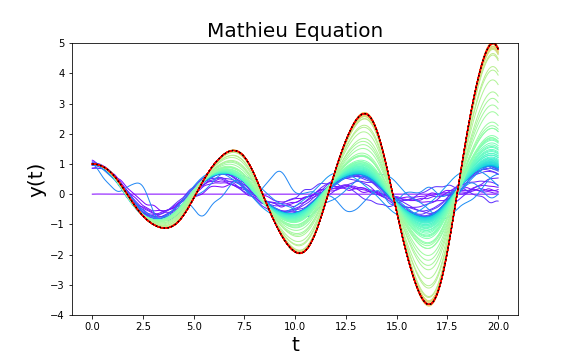}\, 
  \caption{Visualization of the evolution of the solution as represented by the neural network of the double frequency equation, Eq.~\ref{eq:DoubleFreq}, and the Mathieu equation, Eq.~\ref{eq:Mathieu}, left and right figure respectively. The number of training iterations runs from purple (a few) to red (all of them). }
  \label{fig:colorplot}
  \end{center}
\end{figure}

In Figure~\ref{fig:colorplot} we illustrate the evolution of the neural network solutions of the Mathieu equation, see Eq.~\ref{eq:Mathieu}, and the double frequency equation, see Eq.~\ref{eq:DoubleFreq}, as function of number of epoch. In case of the double frequency equation we note that the neural network learns the low base frequency and the high frequency modulations simultaneously. This has to be contrasted with the expected behaviour of neural network interpolation, which typically learns from low to high frequencies~\cite{xu2019training}. However, since the loss function of \textsf{dNNsolve} is sensitive to the derivatives of the learned function as well, which are more enhanced around the high frequency modulations, the observed behaviour makes sense. The same observation was made in~\cite{lu2020deepxde}, although in their example the corresponding differential equation contained the range of frequencies explicitly as a force term. In case of the Mathieu equation we see that the learned solution remains close to a cosine for quite some training iterations until the BFGS optimizer manages to perturb it enough to quickly converge to a better approximation.

\begin{table}
\begin{center}
\renewcommand{\arraystretch}{1.5}
\begin{tabular}{ |p{1cm}|p{1.3cm}|p{1.3cm}|p{1.0cm}|p{1.0cm}|p{1.0cm}|p{1.0cm}|  }
\hline
\multicolumn{7}{|c|}{1D results} \\
\hline
\multicolumn{3}{|c|}{}&\multicolumn{4}{|c|}{$\log_{10}$}\\
\hline
ODE    & Time(s)& Epochs &  $\mc{L}$ &$\mc{L}_{\Omega}$ &$\mc{L}_0$  & $r$ \\
\hline
\ref{eq:Mathieu}      &  16.2   & 1283  & -3.6  & -3.6  & -6.6    & -3.6   (*)   \\
\hline
\ref{eq:Exponential}  &  6.3    & 393   & -4.1  & -4.1  & $<$-25  & -4.1   \\
\hline
\ref{eq:HO}           &  10.6   & 728   & -5.0  & -5.0  & -7.2    & -5.8  \\
\hline
\ref{eq:DHO}          &  11.7   & 765   & -4.6  & -4.6  & -7.2     & -5.1  \\
\hline 
\ref{eq:Linear}       &  6.8    & 513   & -4.2  & -4.2  & $<$-25  & -3.1  \\
\hline
\ref{eq:Delay}        &  7.1    & 414   & -3.2  & -3.2  & $<$-25  &  -2.5   (*) \\
\hline
\ref{eq:Stiff}        &  7.4    & 642   & -3.2  & -3.2  & $<$-25  & -4.3 \\
\hline
\ref{eq:Gaussian}     &  5.5   &  380   & -3.5  & -3.5  & $<$-25  & -3.7  \\
\hline
\ref{eq:DoubleFreq}   &  10.0  &  622   & -3.9  & -3.9  & -6.2    & -4.2  \\
\hline
\ref{eq:Oscillon}     &  7.4   &  74    & -4.9  & -4.9  & -6.8    & -5.2  \\
\hline 
\end{tabular}
\caption{ODE results, entries marked with an asterisk are computed comparing our results with SciPy 1.5.0 \textsf{odeint} solver and \textsf{ddeint} 0.2 solver with $20.000$ points. \label{tab:1D}}
\end{center}
\end{table}

\subsection{2D PDE}
\label{sec:2DPDEs}

In 2D, we considered several boundary value problems on the domain $(x,y) \in [0,1]^2$ with both Dirichlet and Neumann boundary conditions, as well as initial/boundary value problems on the domain $(t, x) \in [0, 1]^2$. We refer to App.~\ref{app:2D} for the explicit equations under study. In the 2D case we used $\{ n_{\Omega}, n_{\partial\Omega}, n_{0}\}=\{1000, 200, 200\}$ collocation points and $n = 10$ neurons per branch. Furthermore, it was possible to successfully solve all the equations reported in App.~\ref{app:2D} by fixing the loss weights to $\{\alpha_0,\alpha_{\partial \Omega}\}=\{10,1\}$. The number of epochs using ADAM optimizer is $210$, while the total number of epochs, including the ones that use BFGS is reported in Tab.~\ref{tab:2D}.

\begin{table}
\begin{center}
\renewcommand{\arraystretch}{1.5}
\begin{tabular}{ |p{1cm}|p{1.3cm}|p{1.3cm}|p{1.0cm}|p{1.0cm}|p{1.0cm}|p{1.0cm}|p{1.0cm}|  }
\hline
\multicolumn{8}{|c|}{2D results} \\
\hline
\multicolumn{3}{|c|}{}&\multicolumn{5}{|c|}{$\log_{10}$}\\
\hline
PDE    & Time(s)& Epochs &  $\mc{L}$ &$\mc{L}_{\Omega}$ &$\mc{L}_0$ &$\mc{L}_{\partial\Omega}$ & $r$ \\
\hline
\ref{eq:Wave2D}            &    8.2   &   528   &  -4.5  & -4.6  & -6.4  & -5.3  &  -5.3 \\
\hline
\ref{eq:Wave2D2}           &    7.5   &   538   &  -4.9  & -4.6  & -6.4  & -5.7  &  -6.3 \\
\hline
\ref{eq:Wave2DTravelling}  &    5.1   &   496   &  -5.1  & -5.3  & -7.1  & -6.1  &  -6.1 \\
\hline
\ref{eq:Heat2D1}           &    8.6   &   911   &  -3.3  & -3.4  & -6.8  & -5.1  &  -4.6  \\
\hline
\ref{eq:Heat2D2}           &    24.8  &   3764  &  -2.8  & -2.8  & -5.0  & -4.0  &  -3.9 \\
\hline
\ref{eq:Heat2D3}           &    8.1   &   762   &  -3.2  & -3.2  & -6.8  & -5.3  &  -4.5 \\
\hline
\ref{eq:Poisson2D1}        &    7.5   &   533   &  -6.4  & -5.2  & -6.2  & -6.3  &  -6.4 \\
\hline
\ref{eq:Poisson2D2}        &    13.4  &   1437  &  -3.5  & -3.6  & -5.5  & -4.4  &  -4.8 \\
\hline
\ref{eq:Advection2D}       &    9.3   &   942   &  -4.2  & -4.3  & -7.0  & -4.9  &  -5.2 \\
\hline
\ref{eq:Burgers2D}         &    22.9  &   3744  &  -2.7  & -3.5  & -3.9  & -3.5  &  -3.9 (*) \\
\hline
\ref{eq:Parabolic2D}       &    9.5   &   1158  &  -3.6  & -3.6  & N/A   & -5.2  &  -5.4 \\
\hline 
\ref{eq:Poisson2DDisk}     &    28.7  &   4574  &  -3.3  & -3.5  &   N/A    & -3.5  &  -5.6 (*) \\
\hline 
\end{tabular}
\caption{2D PDEs results,  entries marked with an asterisk are computed comparing our results with \textsf{Wolfram Mathematica NDSolve} results.\label{tab:2D}}
\end{center}
\end{table}

\subsection{3D PDE}
\label{sec:3DPDEs}

In the 3D case we considered several boundary value problems in the domain $(x, y, z) \in [0,1]^3$, as well as initial/boundary value problems in the domain $(t,x,y) \in [0,1]^3$. However, we need to distinguish the specifics of the solution case by case. For all the 3D PDEs listed in App.~\ref{app:3D} we used $\{ n_{\Omega},n_{\partial\Omega},n_{0}\}=\{ 1000,1200,500\}$ collocation points and we considered $210$ ADAM epochs. 
For the vorticity equation and the Lamb-Oseen vortex we used $n = 20$ neurons per branch, in all other cases $n = 10$ .
The total number of epochs, including the ones that use BFGS is reported in Tab.~\ref{tab:3D}. Moreover, the loss weights needed to be adjusted as described in Tab.~\ref{tab:3D} in order to obtain satisfactory accuracies. \\

\begin{table}
\begin{center}
\renewcommand{\arraystretch}{1.5}
\begin{tabular}{ |p{1cm}|p{1.3cm}|p{1.3cm}|p{1.3cm}|p{1.0cm}|p{1.0cm}|p{1.0cm}|p{1.0cm}|p{1.0cm}|  }
\hline
\multicolumn{9}{|c|}{3D results} \\
\hline
\multicolumn{4}{|c|}{}&\multicolumn{5}{|c|}{$\log_{10}$}\\
\hline
PDE    & Time(s)& Epochs & $(\alpha_0,\alpha_{\partial \Omega})$ & $\mc{L}$ &$\mc{L}_{\Omega}$ &$\mc{L}_0$ &$\mc{L}_{\partial\Omega}$ & $r$ \\
\hline
\ref{eq:Wave3D}            &  22.6   & 706  & (10,1)  & -5-0  & -5.5  & -6.3  & -5.6  & -5.8  \\
\hline
\ref{eq:Wave3D2}           &  27.5   & 524  & (10,1)  & -4.4  & -4.6  & -5.9  & -5.7  & -5.8  \\
\hline
\ref{eq:Wave3DTravelling}  &  15.5   & 715  & (1,1)   & -3.7  & -3.9  & -4.4  & -4.4  & -4.5  \\
\hline
\ref{eq:Heat2D1}           &  24.1   & 750  & (10,10) & -4.0  & -4.7  & -4.5  & -4.3  & -4.5  \\
\hline
\ref{eq:Heat2D2}           &  29.0   & 1484 & (10,10) & -2.9  & -2.9  & -5.0  & -4.7  & -4.8  \\
\hline
\ref{eq:Poisson3D1}        &  30.0   & 1546 & (1,1)   & -2.9  & -3.2  & -3.8  & -3.4  & -3.7  \\
\hline
\ref{eq:Poisson3D2}        &  40.2   & 3277 & (1,1)   & -1.27 & -1.5  & -3.7  & -2.8  & -2.8  \\
\hline
\ref{eq:Poisson3D3}        &  26.1   & 1640  & (1,1)  & -1.5  & -1.8  & -4.0  & -2.8  & -3.0  \\
\hline
\ref{eq:TaylorGreen}       &  48.9   & 1165  & (10,1) & -2.9  & -3.0  & -5.3  & -4.2  & -4.2  \\
\hline
\ref{eq:LambOseen}         &  271.0  & 7389  & (1,1)  & -1.5  & -1.6  & -3.5  & -2.4  & -2.4  \\
\hline
\ref{eq:Vorticity}         &  335.5  & 11232 & (1,1)  & -1.8  & -1.9  & -3.6  & -2.5  & (*) \\
\hline 
\end{tabular}
\caption{PDEs results in 3D. We do not have a solution to compare with for the case marked with an asterisk, the vorticity equation in Eq.~\ref{eq:Vorticity}.
\label{tab:3D}}
\end{center}
\end{table}

\section{Discussion and Conclusions}
\label{sec:Conclusions}

In this paper, we presented a new implementation of the PINN concept to solve ODEs/PDEs using NNs - named \textsf{dNNsolve}. The main novelties of our implementation reside in two points: \textit{i)} the usage of two types of activation functions - sines and sigmoids - as a more efficient basis to expand the solution and \textit{ii)} the use of a new architecture, tailored in a way to provide the desired expansion in terms of this enlarged basis.

It is well known that a Fourier series can be used to expand any continuous function defined on a compact support, which is the case of the ODEs and PDEs considered in this paper. However, as a generic solution will not be just a periodic function, we expect that a more efficient representation than a bare Fourier expansion can be found by including sigmoid activation functions in the expansion basis. We naively expect that the sigmoids might help the network to capture any secular behaviour featured by the solution, while the sines would take care of any periodic behaviour. We also took into account the possible non-linear interplay between a periodic and a non-periodic expansion of the solution using element-wise multiplication of the sine and sigmoid branch. In order to achieve the expansion in this enlarged basis, we introduced a specific architecture, which is pictorially shown in Figure~\ref{fig:architecture}. We call this new implementation \textsf{dNNsolve}: \emph{d}ual \emph{N}eural \emph{N}etwork solve(r) of ODEs and PDEs.

We empirically show that the our expectation is met and the network is able to learn an efficient representation of the solution that most of the times makes use of both the sigmoids and the sines. We also show that in those cases that obviously require only one kind of basis functions (i.e. only sines, sigmoids or multiplication of them), the network is able to adjust the weights accordingly, basically cutting out the redundant branches. On the other hand, given that unused neurons are never exactly set to zero activation, these can bring noise to the solution, compromising the final accuracy. We saw this happening in the oscillon ODE where we had to reduce the number of neurons per branch. This remains an open problem of our architecture, the solution of which we leave for future works.

\textsf{dNNsolve} is able to solve a broad range of ODEs in 1D, see App.~\ref{app:1D}, including both first and second order initial value problems. In particular, it successfully solves the damped harmonic oscillator, whose solution rapidly goes to vanishing amplitude, as well as stiff and delayed equations. The accuracies of these solutions, listed in Tab.~\ref{tab:1D} were computed by comparing with the analytical solutions -- where these are available -- or with the solutions obtained through a standard solver such as the \textsf{odeint} and \textsf{ddeint} packages in Python, or \textsf{NDSolve} in \textsf{Mathematica}. All the root mean squared accuracies are below $10^{-3}$, reaching in 1D values $< 10^{-5}$ for the harmonic oscillator, its damped version and the oscillon profile equation. Interestingly, we were able to solve the oscillon profile equation, see Eq.~\ref{eq:Oscillon}, that is a boundary value problem in 1D, typically solved using the shooting method. We note that tunnelling profiles in phase transitions have been solved with a simple one layer PINN before in~\cite{Piscopo:2019txs}. Even though bubble profiles are expected to be particularly well expanded in a sigmoid base, we find that our NN oscillon solution makes use of the Fourier basis, and becomes much more accurate than the typical root mean squared accuracies $\sim 10^{-3}$ reported in~\cite{Piscopo:2019txs}. We plan to expand our study of these kinds of problems in a separate publication.

In 2D, \textsf{dNNsolve} was used to solve a variety of PDEs, see App.~\ref{app:2D}, including parabolic, hyperbolic and elliptic types on a rectangular domain $[0,1]^2$. All the accuracies are below $10^{-4}$ (except for the heat equation, Eq.~\ref{eq:Heat2D2}), reaching values as small as $10^{-6}$, see Tab.~\ref{tab:2D}. Similar results are obtained in the 3D case, see Tab.~\ref{tab:3D}, in which case all the equations are solved on a cube of the form $[0,1]^3$.

As shown in App.~\ref{sec:AppendixA}, the use of sine activation functions may introduce local minima in the loss function, making it hard for the NN to converge to the true solution of the ODE/PDE. We addressed this problem by introducing two additional hyperparameters $\alpha_0$ and $\alpha_{\partial\Omega}$ that assign different weights to the various parts of the loss function (for initial/boundary value problems), see Eq.~\eqref{eq:Loss}, in the 2D and 3D cases. In particular, in the 2D case we had to weigh the initial conditions part of the loss, $\mathcal{L}_0$, more than the other two pieces in order for \textsf{dNNsolve} to converge to the true solution for all the PDEs considered. In the 3D case it was not possible to find a common choice of the hyperparameters $\alpha$, hence the optimal values are reported in Tab.~\ref{tab:3D}. Overall, we did not fine-tune the hyperparameters $\alpha$ in any of the examples presented: even in the 3D case we just picked one combination of $\alpha_0$ and $\alpha_{\partial_\Omega}$ (choosing their value to be either $1$ or $10$) that led to an acceptable accuracy. We consider this to be one of the main achievements of our work: \textsf{dNNsolve} is able to solve a broad class of differential equations without the need to tune (or with a mild tuning in the 3D case) any hyperparameter. We leave to future work the implementation of an automatic procedure for the optimal choice of the hyperparameters $\alpha$. Also, note that the results we have reported are not the best results achievable with \textsf{dNNsolve}, as they are obtained using a random set of initialized weights as opposed to the best solution selected out of multiple runs~\cite{lu2020deepxde}.

Finally, one interesting question to ask from the physicist's point of view is whether these kinds of solvers might be able to replace standard lattice codes for instance to perform cosmological simulations. It has already been shown~\cite{avrutskiy2020neural} that NN solvers are able to catch up with finite difference methods both in terms of memory complexity (in 3D) and in terms of time complexity (in 5D). Our method is a further step in the direction of making NN-based PDE solvers competitive with standard methods, since it improves the efficiency of the representation of the PDE solution. In order to become fully competitive though, further improvements need to be implemented, among which: \textit{i)} we need to find a way to reduce the noise coming from superfluous neurons, see discussion above; \textit{ii)} we need a better way for the network to avoid local minima, for instance implementing an automatic tuning of the hyperparameters $\alpha$; \textit{iii)} we need an efficient way to parallelize the computation over multiple CPUs and/or GPUs. We plan to come back to these points in future works.

%\section*{Broader impact}

\section*{Acknowlegdments}

AW and YW are supported by the ERC Consolidator Grant STRINGFLATION under the HORIZON 2020 grant agreement no. 647995. F.M. thanks Ben Mares for inspiring discussions about topics related to this paper. F.M. is funded by a UKRI/EPSRC Stephen Hawking fellowship, grant reference EP/T017279/1. This work has been partially supported by STFC consolidated grant ST/P000681/1.

\bibliographystyle{JHEP}
\bibliography{references}

\appendix
\section{Comments on $\alpha$ hyperparameters}
\label{sec:AppendixA}
\begin{figure}[t!]
\begin{center}
  \includegraphics[width=0.3 \textwidth]{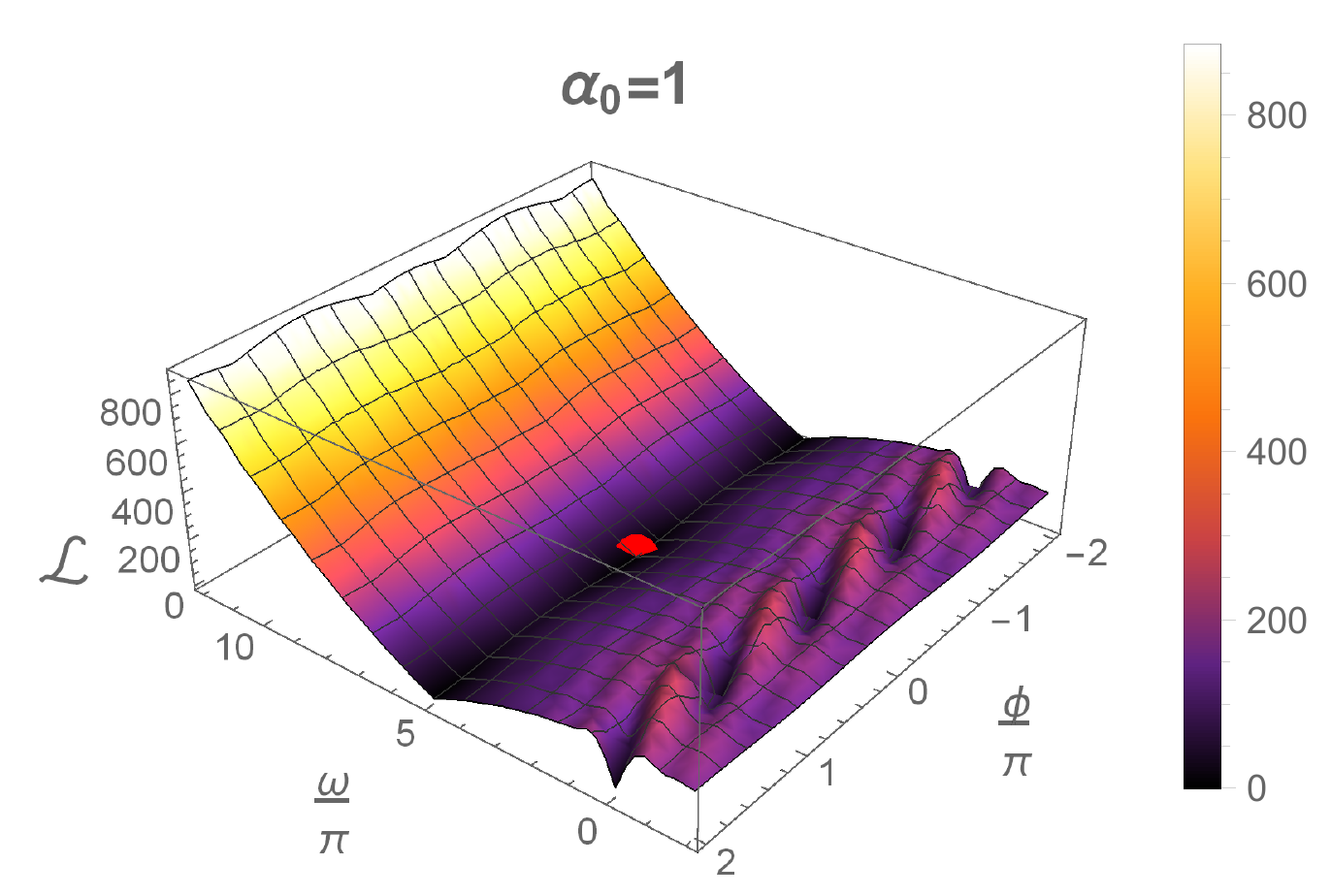}\, \includegraphics[width=0.3 \textwidth]{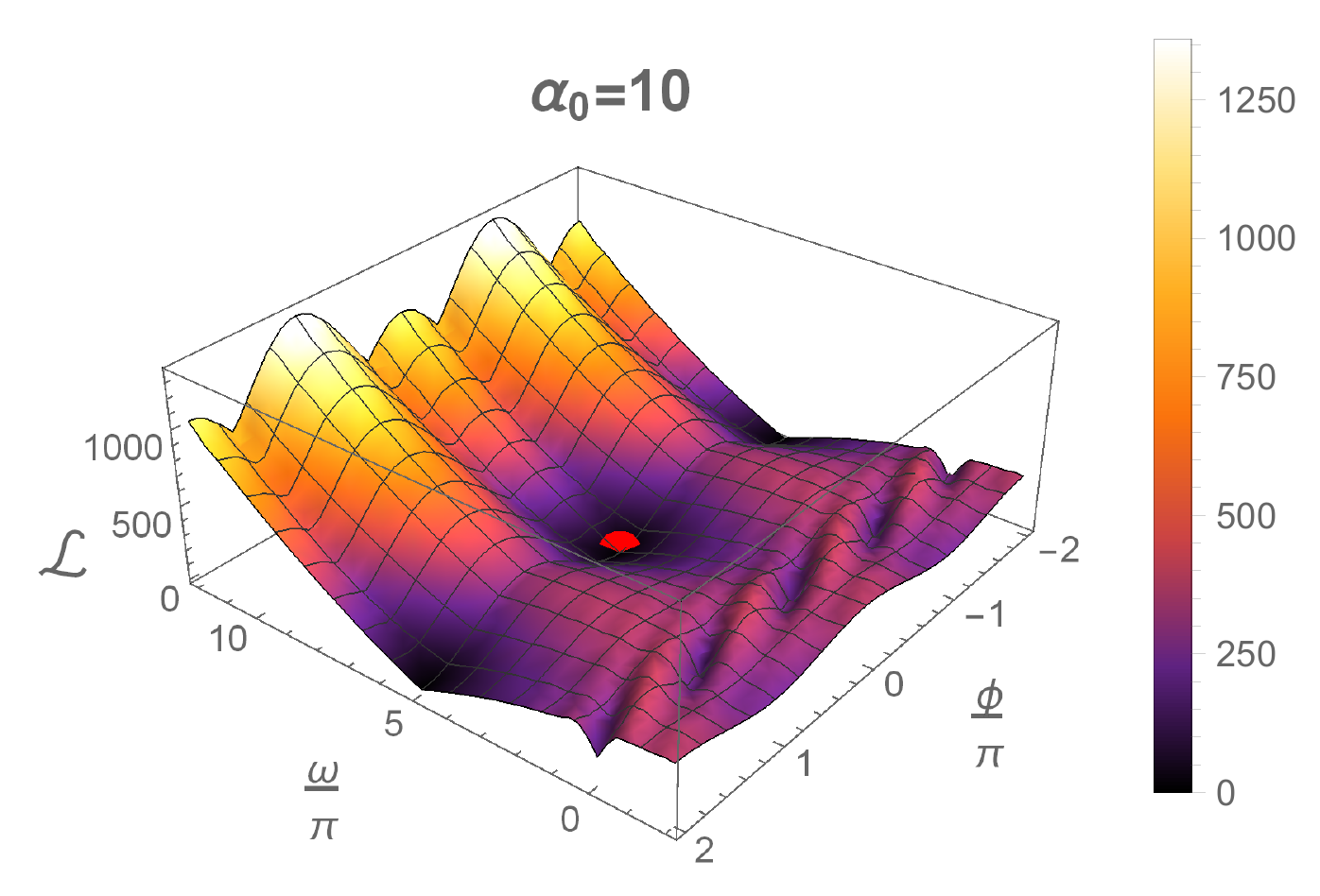}\, \includegraphics[width=0.3 \textwidth]{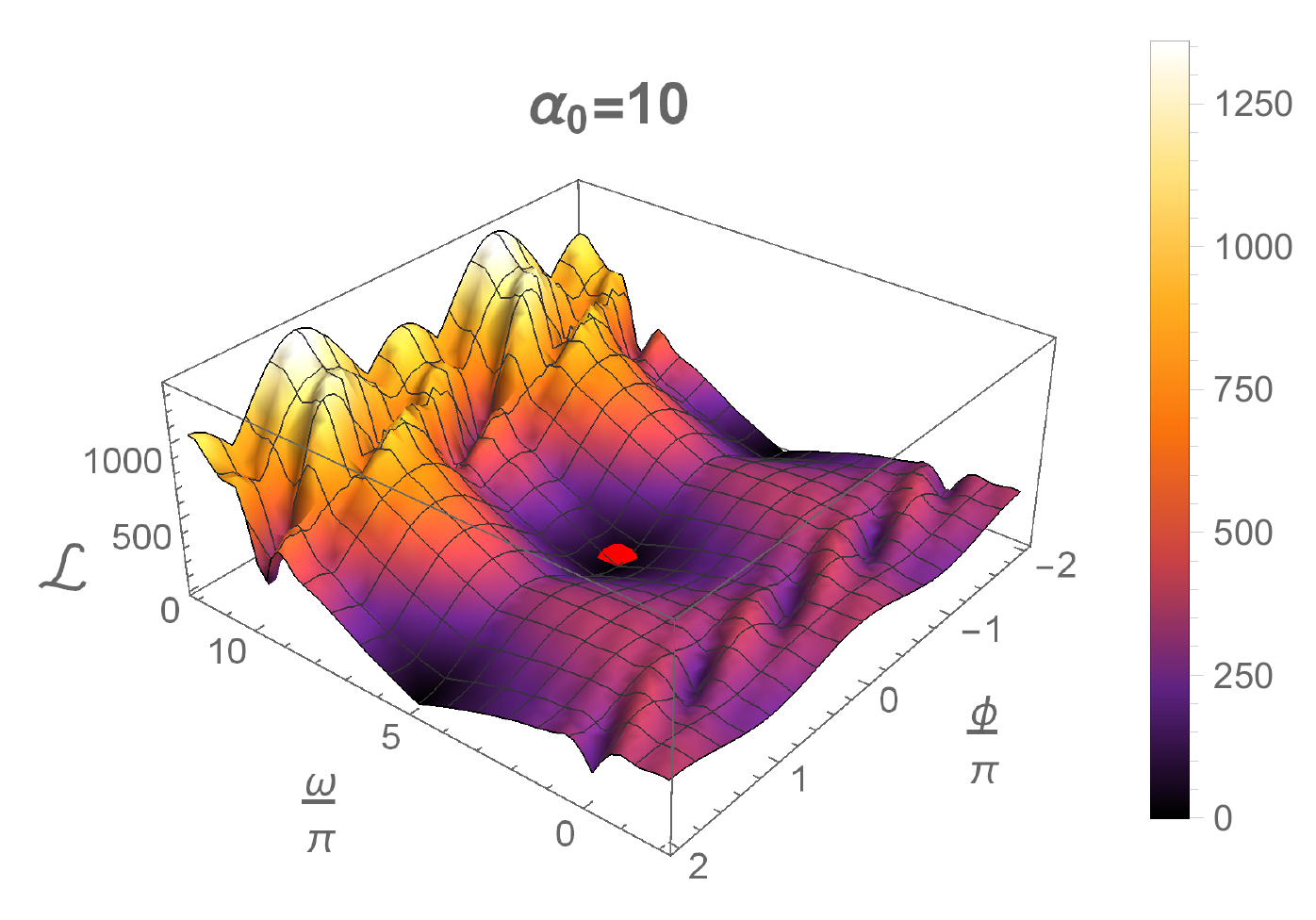}
  \caption{2D loss function $\mc{L}(\omega,\phi)$ obtained from sampling bulk points on a grid of step $\Delta t=0.05$ (left, center) and $\Delta t=0.1$ (right). Left and central plots show the shape of the loss surface when $\alpha_0=1$ and $\alpha_0=10$ respectively. These plots do not reveal any appreciable difference with the analytic result coming from the continuous limit. The right plot shows how artificial local minima can arise from poor sampling.}
  \label{fig:sineloss}
  \end{center}
\end{figure}

In this section we provide some empirical evidence about the advantages of using $\alpha$ hyperparameters to give different weights to the various loss function contributions. To do this, we discuss a couple of examples, focusing on the periodic branch of our architecture. We will consider the simplest Fourier NN with 1 hidden layer and 1 neuron per input variable. 
\paragraph{1D harmonic oscillator}
We want to get some intuition about the loss space when using a sine activation function. In order to do that, we consider a NN having a single sine neuron with a 1D input:
\beq
\hat{u}=d\sin(\omega t +\phi)
\eeq
We try to understand what is the shape of the loss function if we look for solutions of the following harmonic oscillator ODE:
\[\left\{ \begin{array}{llllllll}
 u''+ (5\pi)^2 u=0\\
u(0)=0\\
u'(0)=10\pi\\ 
\end{array}
\right. \qquad t\in[0,1] \] \,.
According to our prescriptions, the loss function in the bulk is given by
\beq
\begin{array}{lll}
\mc{L}_{\Omega}^2&=\frac{1}{n_\Omega}\sum_i \left[ \hat{u}''+ (5\pi)^2 \hat{u}\right]^2 \\[10pt]
&\sim \displaystyle \frac{d^2}{(\Delta x)} \left[(5\pi)^2-\omega^2\right]^2 \int dt \sin^2(\omega t +\phi)\\[10pt]
&= \displaystyle \frac{d^2}{(\Delta x)} \left[(5\pi)^2-\omega^2\right]^2 \left(\frac{2 \omega + \sin(2\phi)-\sin[2(\omega+\phi)]}{4 \omega} \right) \,,
\end{array}
\eeq
while the loss for the initial condition is
\beq
\begin{array}{lll}
\mc{L}_{0}^2&=\frac{1}{2 n_0}\sum_j \left(\left[\hat{u}(0)\right]^2+ \left[\hat{u}'(0))-5\pi\right]^2\right)\\[10pt]
&=\frac{1}{2} \left(d^2\sin^2(\phi)+ \left[d\,\omega \cos(\phi)-10\pi\right]^2\right)\,.\\[10pt]
\end{array}
\eeq
The total loss is given by the weighted sum:
\beq
\mc{L}= \mc{L}_{\Omega} + \alpha_0 \mc{L}_{0}\,.
\eeq
In order to see the relation between the different pieces of the loss function and understand the effects of discretization, we provide some plots of $ \mc{L}$ related to this simple example.
Let us first fix the amplitude parameter to its minimum $d=2$  and plot the 2D loss surface in Figure \ref{fig:sineloss}. These plots are obtained using the discrete definition of the loss, for different values of $ \alpha_0=1,10$ and different numbers of bulk points. Without loss of generality,  we sample the bulk points on an equally spaced grid. We can clearly see the presence of local minima near $\omega=0$ coming from the bulk part of the loss. From the central plot we can infer that setting $\alpha_0\gg 1$ sharpens the location of the global minima located in $\omega=5\pi$, $\phi=0+2\pi k$, $k\in\Bbb Z$. Moreover, comparing the results obtained using $\Delta t=0.05$ and $\Delta t=0.1$, we see that choosing a too wide gridstep may lead to the formation of spurious local minima. 
Despite its simplicity, this example is able to highlight the problems related to NNs with periodic activation functions. On top of local minima coming from the discretization procedure, the loss is intrinsically fulfilled with functions having a huge number of local minima, e.g. cardinal sine.

\begin{figure}[t!]
\begin{center}
  \includegraphics[width=0.3 \textwidth]{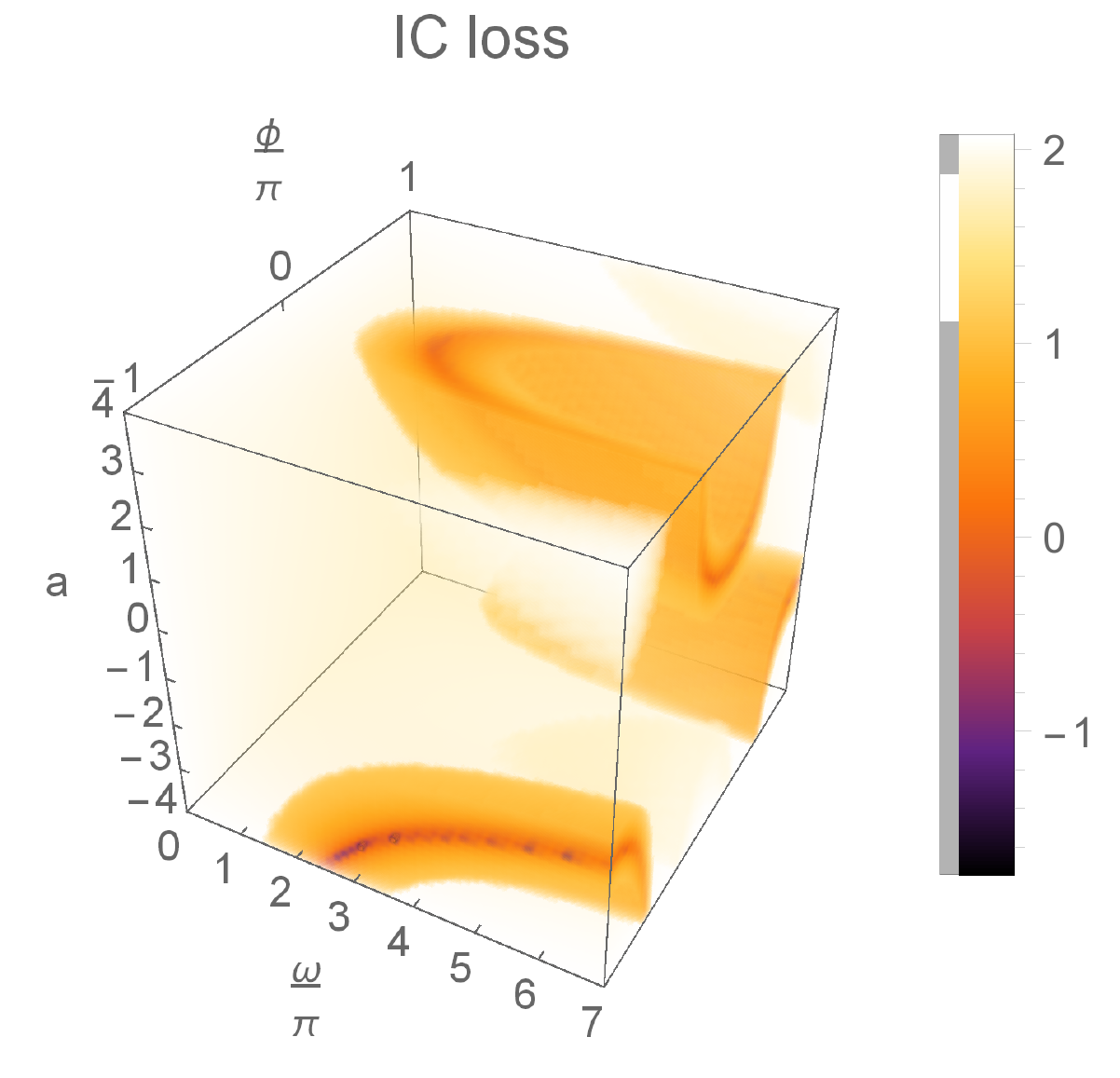}\, \includegraphics[width=0.3 \textwidth]{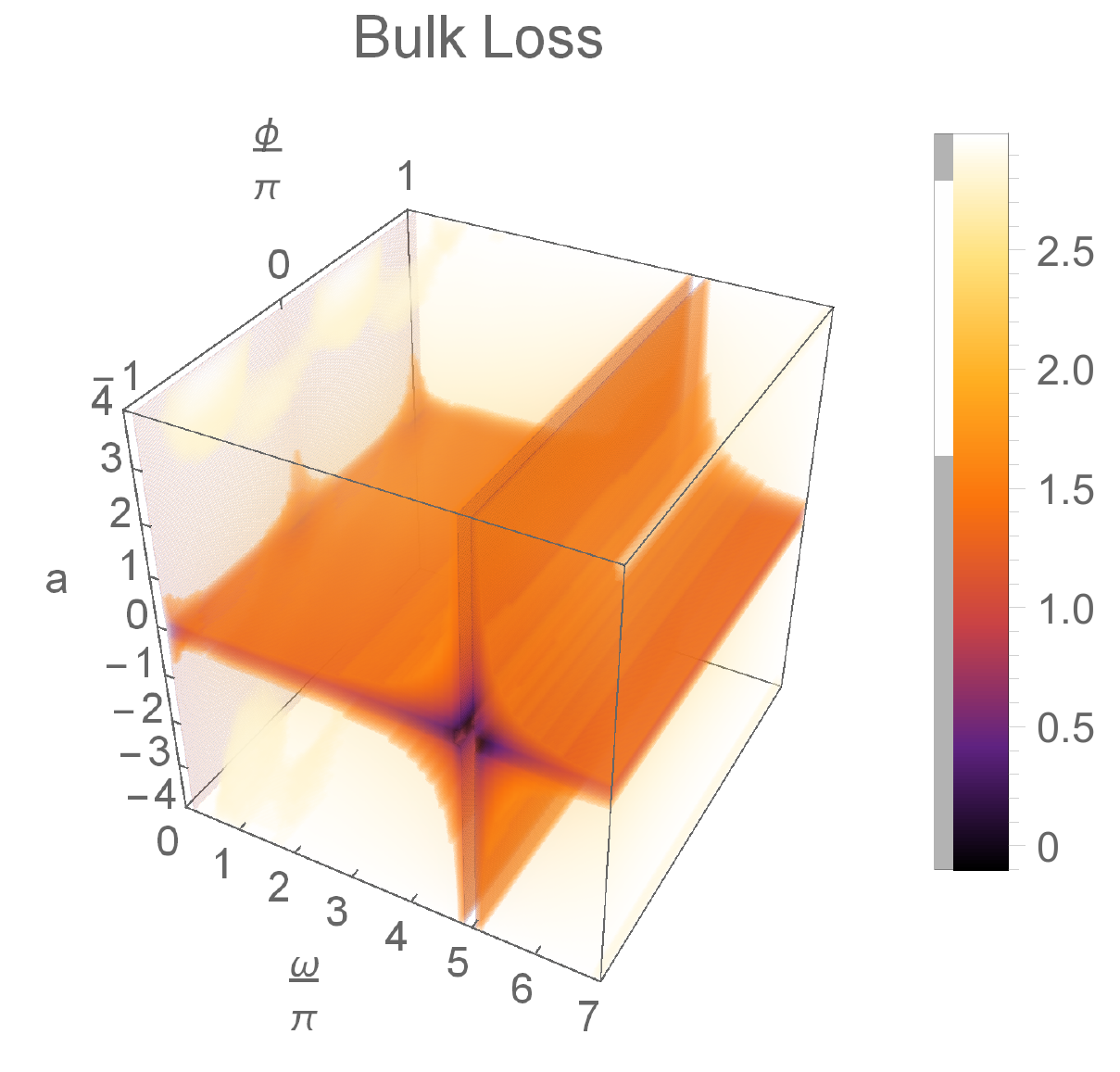}\\[5pt]
   \includegraphics[width=0.3 \textwidth]{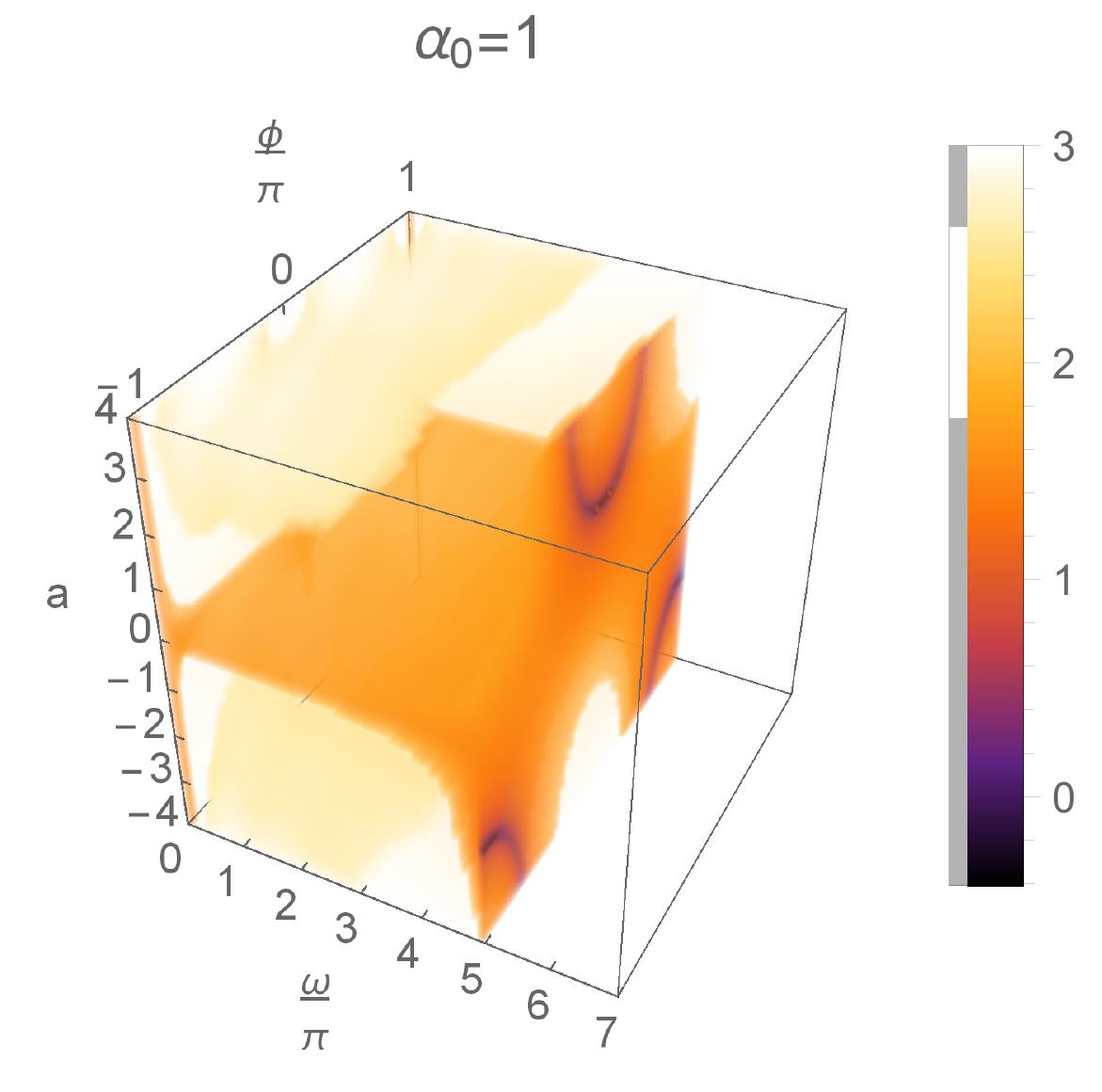}\, \includegraphics[width=0.3 \textwidth]{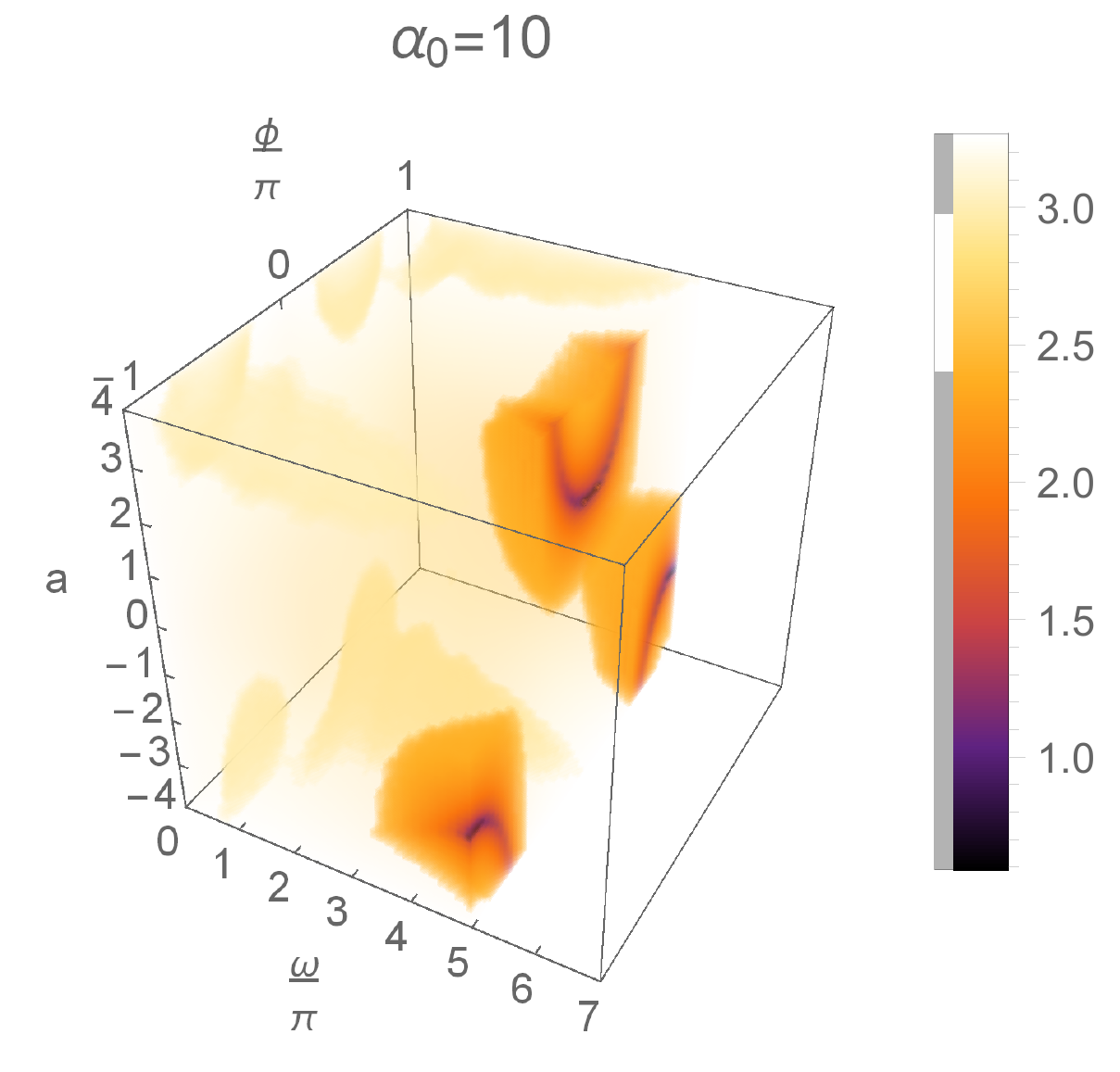}
  \caption{3D loss volume, $\mc{L}(\omega,\phi,d)$, obtained from sampling bulk points on a grid of step $\Delta t=0.1$. Top plots show the IC (left) and bulk contributions (right). Bottom plot show the total loss function computed considering $\alpha_0=1$ (left) and $\alpha_0=10$ (right). To simplify viewing we clip the surface $\omega<5\pi$. These plots do not reveal any appreciable difference with the analytic result coming from the continuous limit.}
  \label{fig:sine_amplitude_loss}
  \end{center}
\end{figure}

Let us now also investigate the r\^ole of the amplitude parameter $d$. The top panels of Figure \ref{fig:sine_amplitude_loss} show the bulk and IC part of the loss function while the total losses, considering $\alpha_0=1,10$, are shown in the bottom panels. As before, these plots are obtained using a grid with $\Delta t=0.1$. From the top plots we can see that IC and bulk loss do not share their local minima. Compared to the previous analysis, the bulk loss exhibits a new local minimum at $d=0$. This comes from the fact that the wave equation is homogeneous and allows for vanishing solutions. Given that $\mc{L}_{\Omega}$ linearly depends on $d$, it is very easy for the network to minimize the $d$ direction. This, coupled to the fact that on average $\mc{L}_{\Omega}>\mc{L}_{0}$, implies that in the early stages of training the network parameters will rapidly reach the $d=0$ surface. If we do not introduce $\alpha$ hyperparameters, this behaviour becomes a serious problem of the training process. Indeed, the bottom-left plot of Figure \ref{fig:sine_amplitude_loss} shows that, choosing $\alpha_0=1$, $d=0$ is a flat local minimum of the loss function, i.e. $\nabla_{\omega,\phi}\mc{L}|_{d=0}=(0,0)$.
On the other hand, according to what we have seen before, setting $\alpha_0\gg 1$ sharpens the location of the global minima (bottom-right plot). This result should not be surprising at all: an ODE defined on a certain domain admits infinite solutions and the result of an ODE problem becomes unique if and only if we provide the initial conditions. Therefore, we must set $\alpha_0\gg 1$ to counteract the loss hierarchy $\mc{L}_{\Omega}>\mc{L}_{0}$.
Indeed, also from a theoretical point of view, the minimization of the bulk loss becomes really informative only if it takes place after or together with the minimization of $\mc{L}_0$.  

\begin{figure}[t!]
\begin{center}
  \includegraphics[width=0.3 \textwidth]{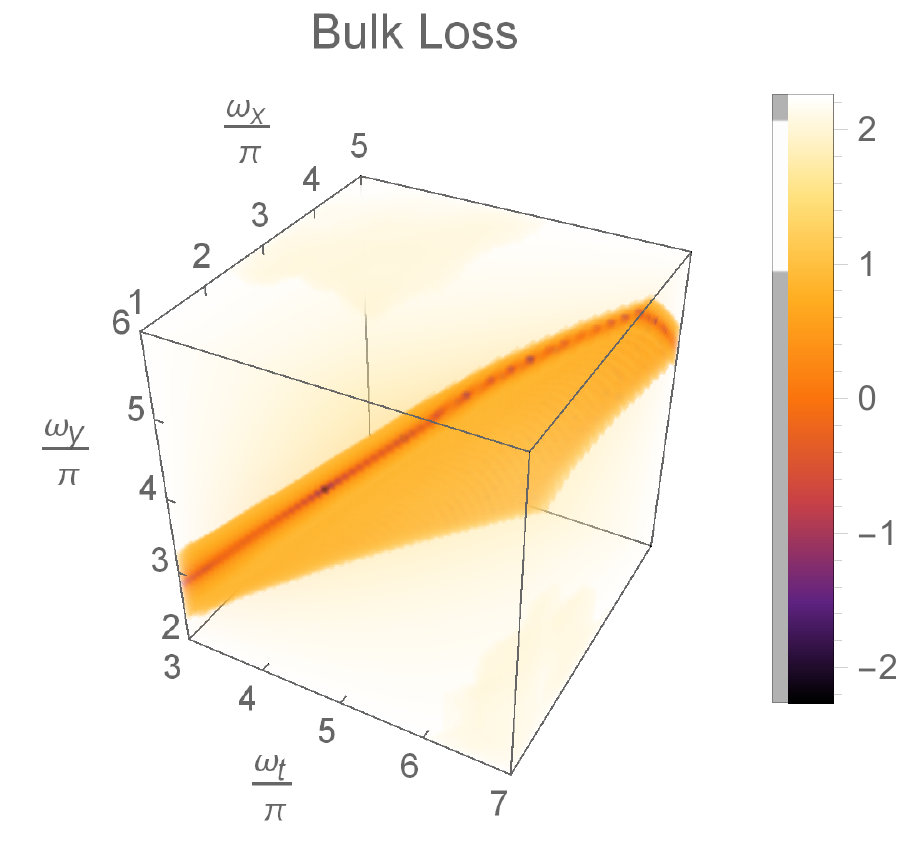}\, \includegraphics[width=0.3 \textwidth]{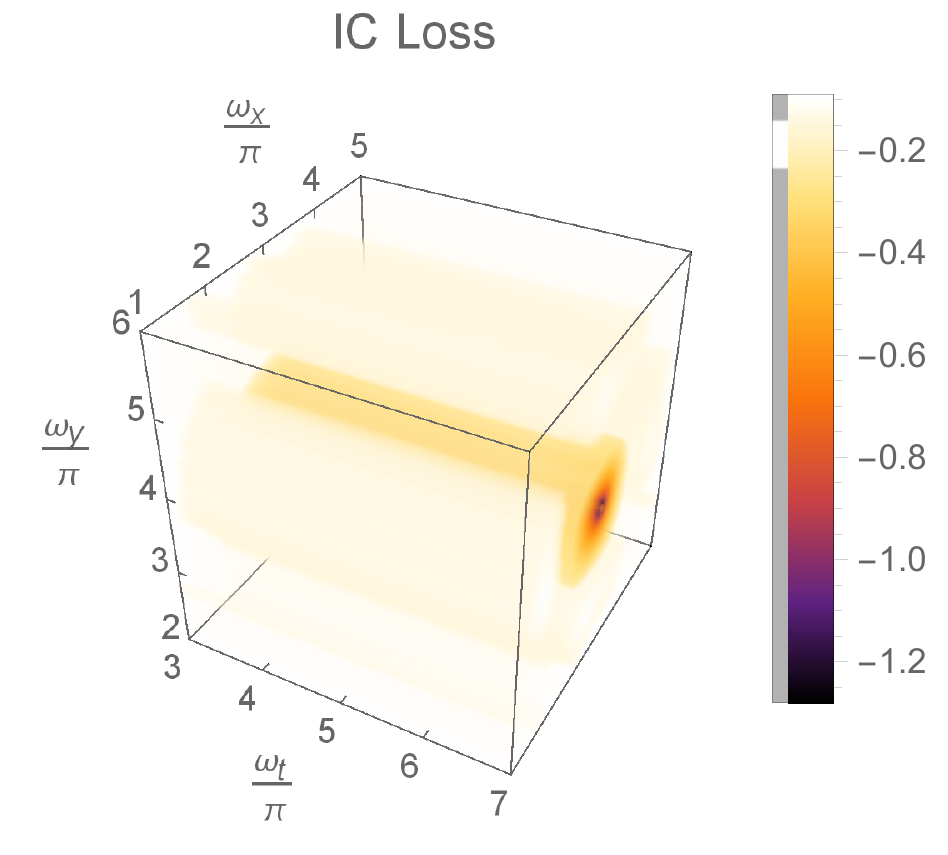}\, \includegraphics[width=0.3 \textwidth]{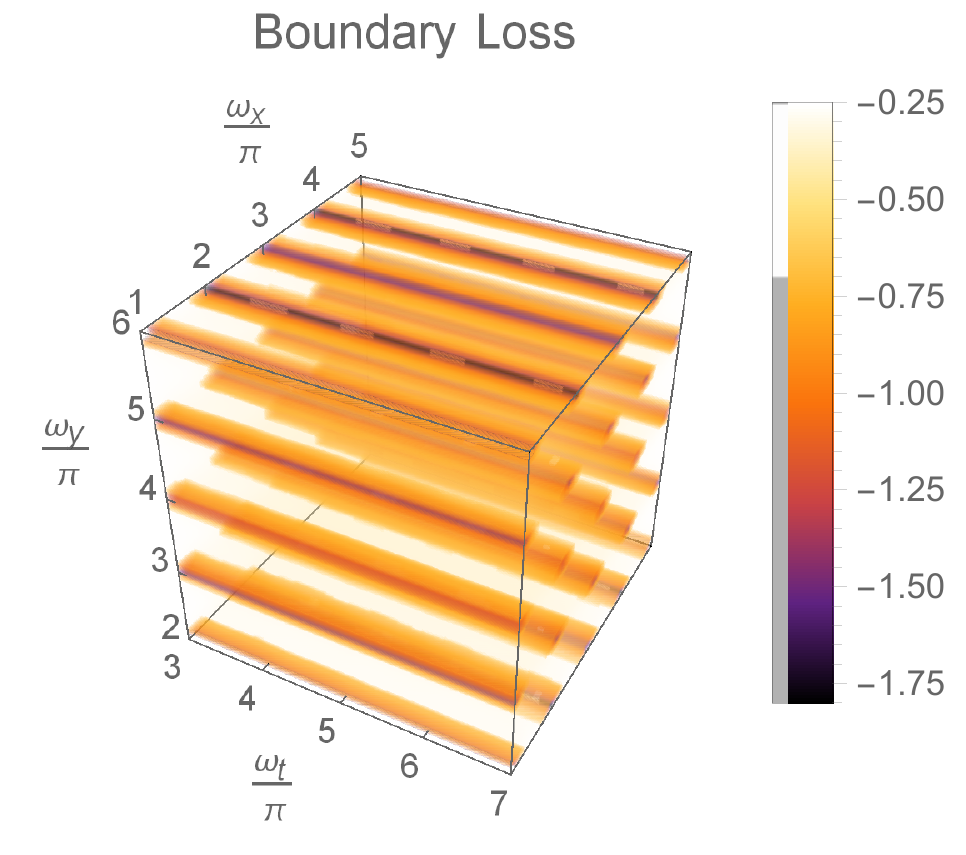}
  \caption{Continuous limit of the different contributions to the loss function setting phase weights to their minima: $\mc{L}_\Omega$ (left), $\mc{L}_0$ (center) and $\mc{L}_{\partial\Omega}$ (right). Discrete versions computed on $11^3$ grid points do not show any significant difference inside this volume. The legend bars refer to $\log_{10}(\mc{L})$.}
\label{fig:lossparts}
\end{center}
\end{figure}

\paragraph{2+1D wave equation}
We focus on a slightly more complex toy model given by solving the 2+1D wave equation (see Eq.~\ref{eq:Wave3D}) with three sine neurons. A simplified version of our sinusoidal branch would be given by:
\beq
\hat{u}=\sin(\omega_t t+\phi_t)\sin(\omega_x x+\phi_x)\sin(\omega_y y+\phi_y) \,.
\eeq
We use this network to solve the following PDE:
\[\left\{ \begin{array}{llllllll}
\partial_{tt}^2 u - (\partial_{xx}^2 u+ \partial_{yy}^2 u) = 0\\
u(0,x,y)=\sin(3 \pi x)\sin(4 \pi y)\\
\partial_t u(0,x,y)=0\\ 
u\left|_{\partial\Omega}\right.=0
\end{array}
\right.  \qquad \Omega=[0,1]^3 \,.\] 
According to the definition given in Eq.~\eqref{eq:Loss}, the parts of the loss function associated to the bulk, the initial condition and the boundary are given by:
\beq
\begin{array}{lll}
\mc{L}_{\Omega}^2&=\displaystyle\frac{1}{n_\Omega}\sum_i \left[
\partial_{tt}^2 \hat{u} - (\partial_{xx}^2 \hat{u}+ \partial_{yy}^2 \hat{u})\right]^2 \\[10pt]
&\sim \displaystyle \frac{1}{V(\Omega)} \displaystyle \int_\Omega dx dy dt \left[
\partial_{tt}^2 \hat{u} - (\partial_{xx}^2 \hat{u}+ \partial_{yy}^2 \hat{u})\right]^2  \\[10pt]
&= \displaystyle\frac{(\omega_x^2+\omega_y^2-\omega_t^2)^2}{V(\Omega)} \displaystyle \int_\Omega dx dy dt \left[\sin(\omega_t t+\phi_t)\sin(\omega_x x+\phi_x)\sin(\omega_y y+\phi_y)\right]^2  \\[10pt]
&= \displaystyle\frac{(\omega_x^2+\omega_y^2-\omega_t^2)^2}{V(\Omega)} \displaystyle \prod_{i=t,x,y} \left(\frac{2 \omega_i + \sin(2\phi_i)-\sin[2(\omega_i+\phi_i)]}{4 \omega_i} \right) \,, \\ 
\end{array}
\eeq

\beq
\begin{array}{lll}
\mc{L}_{0}^2&=\displaystyle\frac{1}{2 n_0}\sum_j \left(\left[\hat{u}(t_0)-\sin(3 \pi x)\sin(4 \pi y)\right]^2+ \left[[\partial_t \hat{u}](t_0))\right]^2\right)\\[10pt]
&\displaystyle\sim\frac{1}{2A(\Omega_0)}\int_{\Omega_0}dx dy \left(\left[\hat{u}(t_0)-\sin(3 \pi x)\sin(4 \pi y)\right]^2+ \left[[\partial_t \hat{u}](t_0))\right]^2\right)\\[10pt]
&\displaystyle=\frac{1}{2A(\Omega_0)}\left[\displaystyle  \int_{\Omega_0}dx dy \left[\sin(\phi_t)\sin(\omega_x x+\phi_x)\sin(\omega_y y+\phi_y)-\sin(3 \pi x)\sin(4 \pi y)\right]^2+\right. \\[10pt]
&\displaystyle \qquad +\left. \int_{\Omega_0}dx dy \left[\omega_t \cos(\phi_t)\sin(\omega_x x+\phi_x)\sin(\omega_y y+\phi_y)\right]^2\right]\\[10pt]
&\displaystyle =\frac{1}{2A(\Omega_0)}\left[\frac{1}{4} -24\pi^2\sin(\phi_t)\left(\frac{\sin(\phi_x)+\sin(\phi_x+\omega_x)}{(9\pi^2-\omega_x^2)}\right)\left(\frac{\sin(\phi_y)+\sin(\phi_y+\omega_y)}{(16\pi^2-\omega_6^2)}\right)\right.\\[10pt]
&\qquad\left.\displaystyle + \left(\sin^2(\phi_t)+\omega_t^2\cos^2(\phi_t)\right)\prod_{i=x,y}\left(\frac{2 \omega_i + \sin(2\phi_i)-\sin[2(\omega_i+\phi_i)]}{4 \omega_i} \right)\right] \,,\\[10pt]
\end{array}
\eeq

\beq
\begin{array}{lll}
\mc{L}_{\partial \Omega}^2&\displaystyle=\frac{1}{n_{\partial\Omega}} \sum_k \left[\hat{u}|_{\partial\Omega} \right]^2\\[10pt]
&\displaystyle\sim \frac{1}{4A(\partial\Omega)}\displaystyle \left( \int_{x=0}dy dt\left[\hat{u} \right]^2+ \int_{x=1}dy dt\left[\hat{u} \right]^2+ \int_{y=0}dy dt\left[\hat{u} \right]^2+ \int_{y=1}dy dt\left[\hat{u} \right]^2\right)\\[10pt]
&\displaystyle= \frac{1}{4A(\partial\Omega)}\displaystyle \left( (\sin^2(\phi_x)+\sin^2(\omega_x +\phi_x)) \int dy dt (\sin^2(\omega_t t+\phi_t)\sin^2(\omega_y y+\phi_y)) \right.\\[10pt]
&\qquad \left.+ (\sin^2(\phi_y)+\sin^2(\omega_y +\phi_y)) \displaystyle \int dy dt (\sin^2(\omega_t t+\phi_t)\sin^2(\omega_x x+\phi_x)\right)\\[10pt]
&\displaystyle= \frac{1}{4A(\partial\Omega)}\displaystyle \left\{\left(\frac{2 \omega_t + \sin(2\phi_t)-\sin[2(\omega_t+\phi_t)]}{4 \omega_t} \right) \times\right.\\[10pt]
&\qquad  \displaystyle \left[ (\sin^2(\phi_x)+\sin^2(\omega_x +\phi_x)) \left(\frac{2 \omega_y + \sin(2\phi_y)-\sin[2(\omega_y+\phi_y)]}{4 \omega_y}\right) \right.\\[10pt]
&\qquad  \displaystyle \left.\left. (\sin^2(\phi_y)+\sin^2(\omega_y +\phi_y)) \left(\frac{2 \omega_x + \sin(2\phi_x)-\sin[2(\omega_x+\phi_x)]}{4 \omega_x}\right)   \right]\right\}\,.
\end{array}
\eeq
We plot $\mc{L}_{\Omega}$, $\mc{L}_{0}$ and $\mc{L}_{\partial\Omega}$  in Figure \ref{fig:lossparts}, fixing the phase parameters at their minima. These results show that there is a natural hierarchy, i.e.  $\mc{L}_{\Omega}\gg\mc{L}_{0},\mc{L}_{\partial\Omega}$, that can be explained by the effect of derivation on Fourier NN: each derivative provides a $\omega$ power in the loss function. This results in a polynomial behaviour of $\mc{L}_{\Omega}$ in the frequency space, while initial conditions and boundary conditions may be at most linear. Given that the initial conditions of the parameters are randomly chosen, this implies that, on average, $\mc{L}_{\Omega}$ will be the major contribution to the loss function in the early stages of training. 

\begin{figure}[t!]
\begin{center}
  \includegraphics[width=0.3 \textwidth]{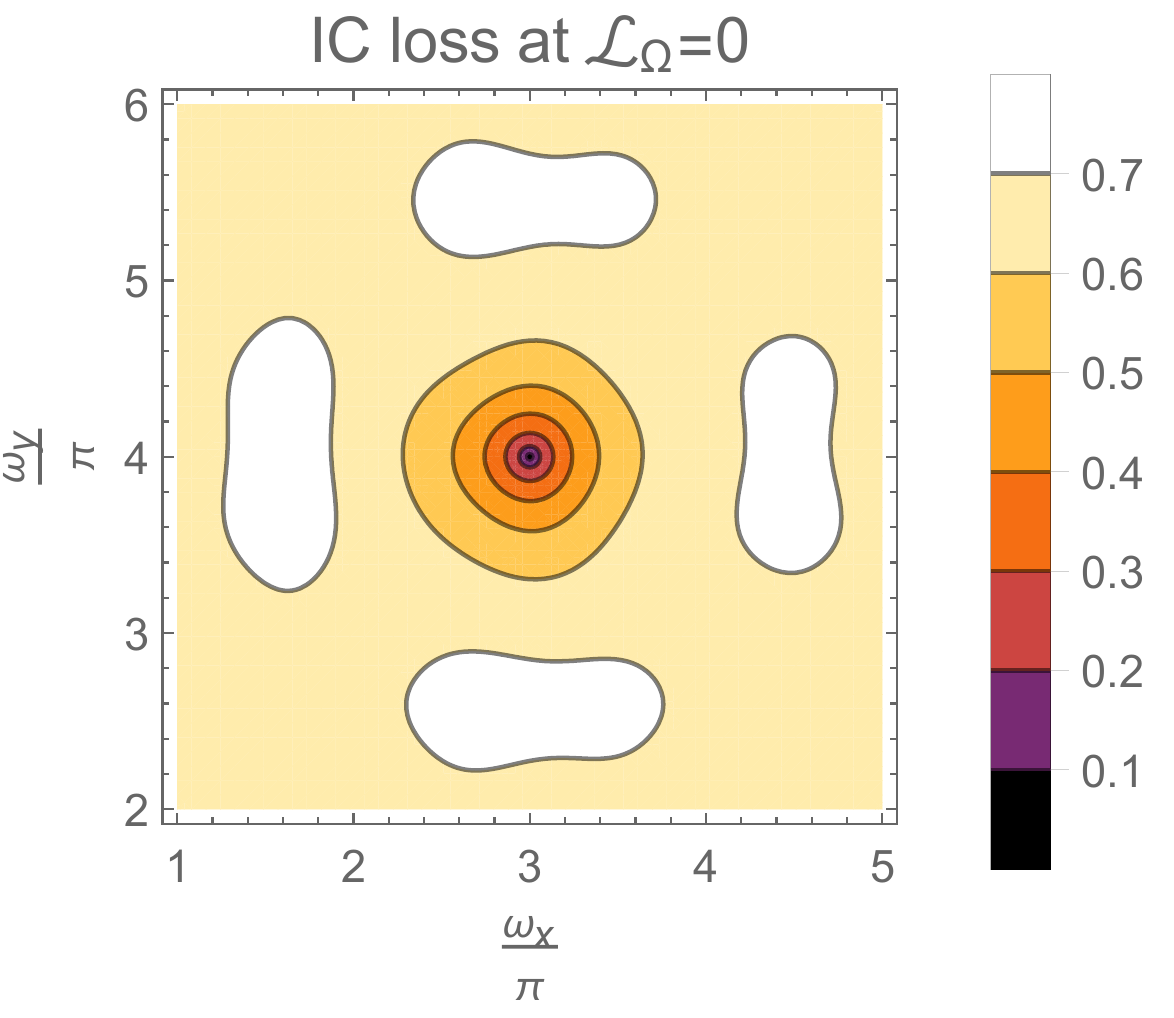}\, \includegraphics[width=0.3 \textwidth]{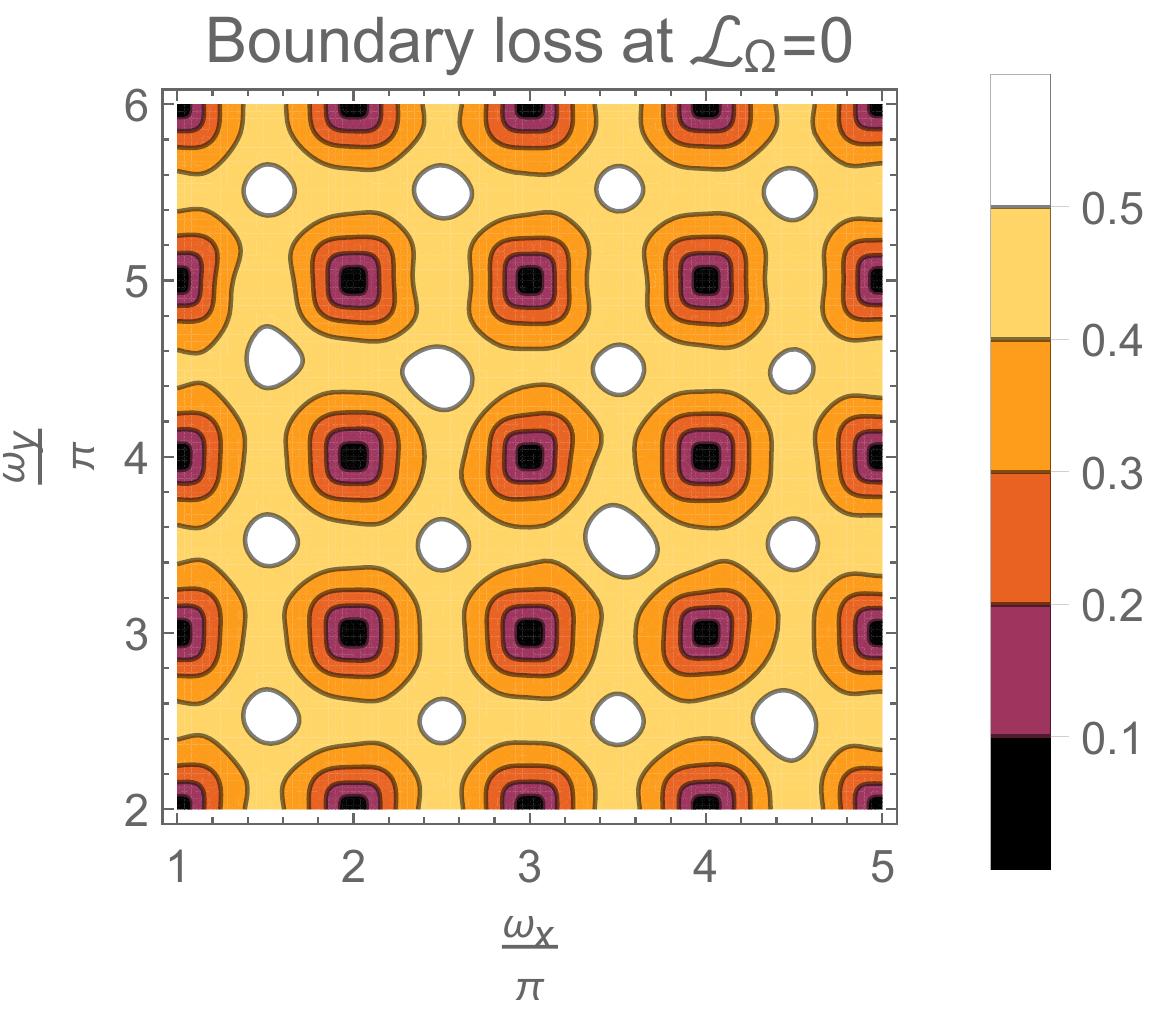}\, \includegraphics[width=0.3 \textwidth]{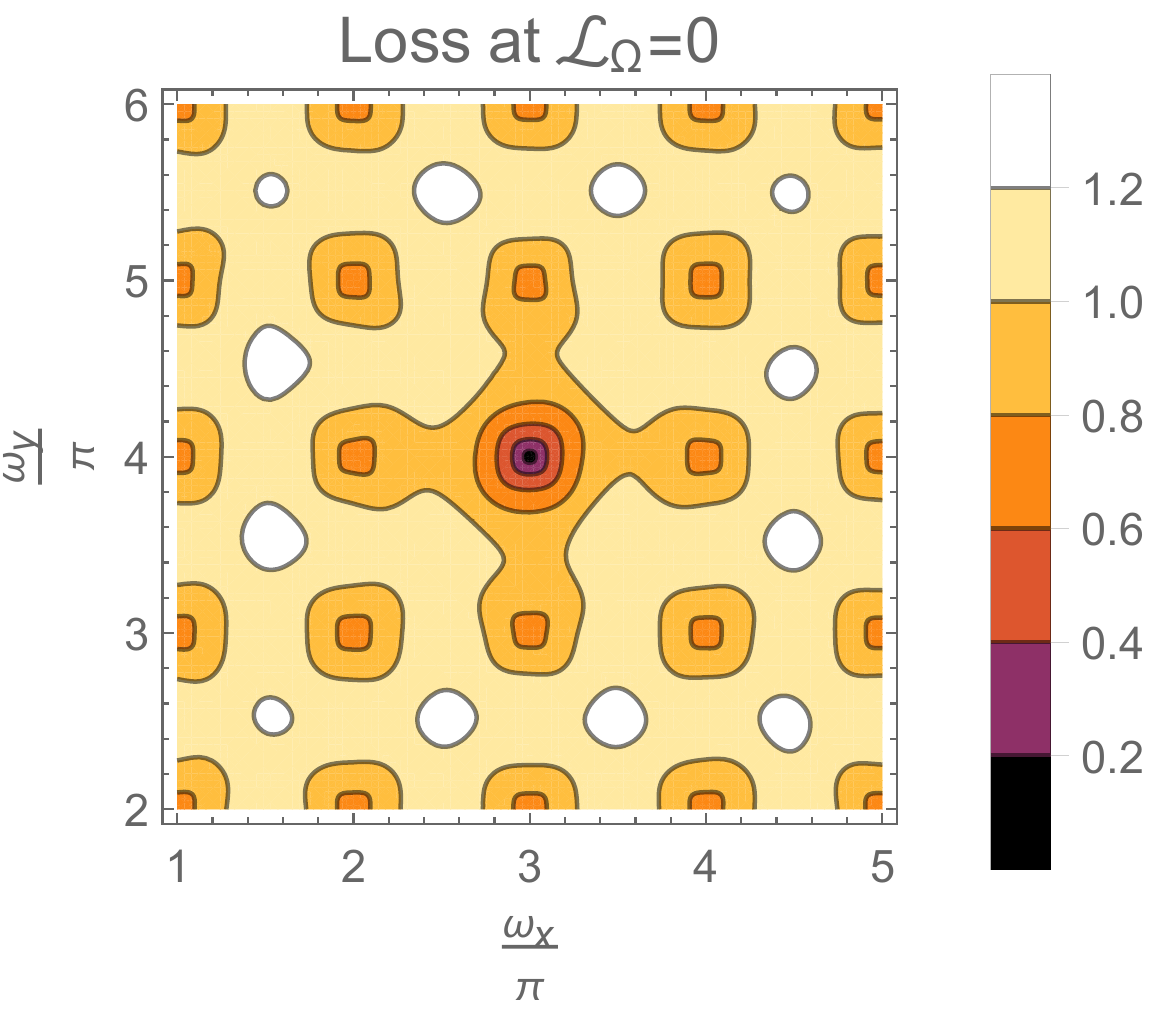}
  \caption{These 1970s typical tiles show the 2D loss function that appears after $\mc{L}_\Omega$ minimization given by $(\omega_x^2+\omega_y^2-\omega_t^2)=0$. We consider a grid of $11^2$ points. While the contribution coming from IC (left) shows a single minimum, the boundary contribution is fulfilled with equally spaced minima (center).  The total residual loss (right) is then characterised by the presence of infinitely many local minima surrounding the true solution. This makes the NN training extremely hard and the system gets usually stuck in a local minimum.  }
\label{fig:minima}
  \end{center}
\end{figure}

Going back to our example, this means that training will lead the parameters toward the surface $(\omega_x^2+\omega_y^2-\omega_t^2)=0$ that is clearly visible in the left plot of Figure \ref{fig:architecture}. From then on, the trajectory in the parameter space will not be able to leave the surface and there is a high probability of getting stuck in one of the minima of $\mc{L}_{\partial\Omega}$, see Figure \ref{fig:minima}. One way to overcome this problem is to impose some hierarchy among the different contributions to the loss function. This can be done by setting the pieces that carry more information about the solution to higher values. In the specific case analyzed in this section, this results in setting $\alpha_0\gg 1$.

\section{1D ODEs under study}
\label{app:1D}
\begin{enumerate}[label={[\arabic*]}]

\item{Mathieu equation:
\[\left\{ \begin{array}{llllllll}
u''(t)+(a-2\,q\, \cos(2 t))u(t)=0 \qquad a=1, q=0.2 \\
u(0)=1\\
u'(0)=0\\
\end{array}
\right.\] \label{eq:Mathieu}}

\item{Decaying exponential:
\[\left\{ \begin{array}{llllllll}
u'(t)+ \beta\, u(t)=0 \qquad \beta=0.52 \\
u(0)=1\\
\end{array}
\right.\]
with analytical solution
\[u(t)= e^{-\beta t}\]\label{eq:Exponential}}

\item{Harmonic oscillator:
\[\left\{ \begin{array}{llllllll}
u''(t)+ \omega^2 u(t)=0 \qquad \omega=5 \\
u(0)=1\\
u'(0)=0\\
\end{array}
\right.\]
with analytical solution
\[u(t) = \cos(\omega\, t)\]\label{eq:HO}}

\item{Damped harmonic oscillator:
\[\left\{ \begin{array}{llllllll}
u''(t)+\beta\, u'(t)+ \omega^2\, u(t)=0\\
u(0)=1\\
u'(0)=0\\
\end{array}
\right.\]
with analytical solution
\[u(t) =e^{-\beta\, t/2}\left( \cos(f\, t)+\frac{\beta}{2f}\sin(f\, t)\right), \quad f\equiv \sqrt{\omega^2-\beta^2/4}\]\label{eq:DHO}}

\item{Linear function:
\[\left\{ \begin{array}{llllllll}
u'(t)-1=0\\
u(0)=1\\
\end{array}
\right.\]
with analytical solution
\[u(t)=1+t\]\label{eq:Linear}}

\item{Delay equation:
\[\left\{ \begin{array}{llllllll}
u'(t)-\beta u(t) + u(t-d)=0, u(t|t<0) = t-1, \qquad d=1\\
u(0)=1\\
\end{array}
\right.\]\label{eq:Delay}}

\item{Stiff equation::
\[\left\{ \begin{array}{llllllll}
u'(t)+21\, u(t) - e^{-t} = 0\\
u(0)=1\\
\end{array}
\right.\]
with analytical solution
\[u(t) =\displaystyle \frac{1}{20}\left(e^{-t}+19e^{-21 t}\right)\]\label{eq:Stiff}}

\item{Gaussian:
\[\left\{ \begin{array}{llllllll}
u'(t)+2\,b\,t\, u(t) =0 \qquad b=0.1\\
u(0)=1\\
\end{array}
\right.\]
with analytical solution
\[u(t) =\displaystyle e^{-b t^2}\]\label{eq:Gaussian}}

\item{Two frequencies:
\[\left\{ \begin{array}{llllllll}
u''(t)+u(t) + A_1\, \cos(\omega_1\, t)+A_2\, \sin(\omega_2\, t), \qquad A_1 = 2, A_2=6, \omega_1=5, \omega_2 = 10\\
u(0)=1\\
u'(0)=0\\
\end{array}
\right.\]
with analytical solution
\[u(t) = \frac{1}{132}(121 \cos(t) + 11 \cos(5t)-80\sin(t)+8\sin(10t)\]\label{eq:DoubleFreq}}

\item{Oscillon Profile equation:
\[\left\{ \begin{array}{llllllll}
u''(t)+\frac{d-1}{r}\,u'(t)+m^2\,u(t)-2\,u^3(t) = 0, \qquad d=1\\
u'+m u =0 \quad \mbox{when}\quad  t\rightarrow \infty\\
u'(0)=0\\
\end{array}
\right.\]
with analytical solution
\[u(t) = \displaystyle \frac{m}{\cosh(m\, t)}\]\label{eq:Oscillon}}

\end{enumerate}

\section{2D PDEs under study}
\label{app:2D}
\begin{enumerate}[label={[\arabic*]}]

\item{Wave equation:
\[\left\{ \begin{array}{llllllll}
\partial_{tt}^2 u - \partial_{xx}^2 u = 0 \\
u(0,x)=\sin(3\pi x)\\
\partial_t u(0,x)=0\\
u\left|_{\partial\Omega}\right.=0
\end{array}
\right.\]
with analytical solution
\[u(t,x)=\cos(3\pi t)\sin(3\pi x)\]
\label{eq:Wave2D}}

\item{Wave equation: 
\[\left\{ \begin{array}{llllllll}
\partial_{tt}^2 u - \partial_{xx}^2 u = 0\\
u(0,x)=\sin(3\pi x)\\
\partial_t u(0,x)=0 \\
\partial_x u\left|_{\partial\Omega}\right.=0
\end{array}
\right.\]
with analytical solution
\[u(t,x)=\cos(3\pi t)\cos(3\pi x)\]\label{eq:Wave2D2}}

\item{Traveling Wave equation:
\[\left\{ \begin{array}{llllllll}
\partial_{t} u - \partial_{x} u = 0 \\
u(0,x)=\sin(2\pi x)\\
u\left|_{\partial\Omega}\right.=\sin(2\pi t)
\end{array}
\right.\]
with analytical solution
\[u(t,x)=\cos\left(2\pi (t+ x)\right)\]\label{eq:Wave2DTravelling}}

\item{Heat equation 1:
\[\left\{ \begin{array}{llllllll}
\partial_{t} u - 0.05\; \partial_{xx}^2 u = 0\\
u(0,x)=\sin(3\pi x)\\
u\left|_{\partial\Omega}\right.=0
\end{array}
\right.\]
with analytical solution
\[u(t,x)=\sin(3\pi x)e^{-0.05(3\pi)^2t}\]\label{eq:Heat2D1}}

\item{Heat equation 2:
\[\left\{ \begin{array}{llllllll}
\partial_{t} u - 0.01\; \partial_{xx}^2 u = 0\\
u(0,x)=2\sin(9\pi x)+0.3 \sin(4\pi x)\\
u\left|_{\partial\Omega}\right.=0
\end{array}
\right.\]
with analytical solution
\[u(t,x)=2 \sin(9\pi x)e^{-0.01(9\pi)^2t}-0.3 \sin(4\pi x)e^{-0.01(4\pi)^2t}\]\label{eq:Heat2D2}}

\item{Heat equation 3: 
\[\left\{ \begin{array}{llllllll}
\partial_{t} u - 0.05\; \partial_{xx}^2 u = 0\\
u(0,x)=\sin(3\pi x)\\
\partial_x u\left|_{\partial\Omega}\right.=0
\end{array}
\right.\]
with analytical solution
\[u(t,x)=\cos(3\pi x)e^{-0.05(3\pi)^2t}\]\label{eq:Heat2D3}}

\item{Poisson equation 1:
\[\left\{ \begin{array}{llllllll}
\partial_{tt}^2 u + \partial_{xx}^2 u + 2\pi^2\sin(\pi t)\sin(\pi x) = 0\\
u\left|_{\partial\Omega}\right.=0
\end{array}
\right. \]
with analytical solution
\[u(t,x)=\sin(\pi x)\sin(\pi t)\]\label{eq:Poisson2D1}}

\item{Poisson equation 2:
\[\left\{ \begin{array}{llllllll}
\partial_{tt}^2 u + \partial_{xx}^2 u +10(t-1)\cos(5x)+25(t-1)(x-1)\sin(5x) = 0\\
u(0,x)=(1-x)\sin(5x)\\
u(1,x)=u(t,0)=u(t,1)=0
\end{array}
\right.\]
with analytical solution
\[u(t,x)=(1-t)(1-x)\sin(5 x)\]\label{eq:Poisson2D2}}

\item{Advenction diffusion equation:
\[\left\{ \begin{array}{llllllll}
\partial_{t} u -\frac{1}{4} \partial_{xx}^2= 0\\
u(0,x)=\frac{1}{4}\sin(\pi x)\\
u(t,0)=u(t,1)=0
\end{array}
\right.\]
with analytical solution
\[u(t,x)=\frac{1}{4}e^{-\frac{1}{4}\pi^2 t}\sin(\pi x)\]\label{eq:Advection2D}}

\item{Burger's equation: 
\[\left\{ \begin{array}{llllllll}
\partial_{t} u +u\partial_x u-\frac{1}{4} \partial_{xx}^2= 0\\
u(0,x)=x(1-x)\\
u(t,0)=u(t,1)=0
\end{array}
\right.\]\label{eq:Burgers2D}}

\item{Parabolic equation on unit disk:
\[\left\{ \begin{array}{llllllll}
\partial_{tt}^2 u + \partial_{xx}^2 u -4= 0\\
u\left|_{\partial\Omega}\right.=1
\end{array}
\right.\]
with analytical solution
\[u(t,x)=\frac{1}{4}e^{-\frac{1}{4}\pi^2 t}\sin(\pi x)\]\label{eq:Parabolic2D}}

\item{Poisson equation on unit disk:
\[\left\{ \begin{array}{llllllll}
\partial_{tt}^2 u + \partial_{xx}^2 u -e^{-(t^2+10x^2)}= 0\\
u\left|_{\partial\Omega}\right.=0
\end{array}
\right.\]\label{eq:Poisson2DDisk}}
\end{enumerate}

\section{3D PDEs under study}
\label{app:3D}
\begin{enumerate}[label={[\arabic*]}]
\item{Wave equation:
\[\left\{ \begin{array}{llllllll}
\partial_{tt}^2 u - (\partial_{xx}^2 u+ \partial_{yy}^2 u) = 0\\
u(0,x,y)=\sin(\pi x)\sin(\pi y)\\
\partial_t u(0,x,y)=0\\
u\left|_{\partial\Omega}\right.=0
\end{array}
\right.\]
with analytical solution
\[u(t,x,y)=\cos(\sqrt{2}\pi t)\sin(\pi x)\sin(\pi y)\]\label{eq:Wave3D}}

\item{Wave equation:
\[\left\{ \begin{array}{llllllll}
\partial_{tt}^2 u - (\partial_{xx}^2 u+ \partial_{yy}^2 u) = 0\\
u(0,x,y)=\sin(3 \pi x)\sin(4 \pi y)\\
\partial_t u(0,x,y)=0\\ 
u\left|_{\partial\Omega}\right.=0
\end{array}
\right.\]
with analytical solution
\[u(t,x,y)=\cos(5\pi t)\sin(3\pi x)\sin(4\pi y)\]\label{eq:Wave3D2}}

\item{Traveling wave equation: 
\[\left\{ \begin{array}{llllllll}
\partial_{t} u -\frac{1}{5}( \partial_{x} u +\partial_{y} u )= 0\\
u(0,x,y)=\sin(3\pi x+2\pi y)\\
\mbox{Dirichlet BC}\\
\end{array}
\right.\]
with analytical solution
\[u(t,x,y)=\sin(3\pi x+ 2\pi y + \pi t)\]\label{eq:Wave3DTravelling}}

\item{Heat equation 1:
\[\left\{ \begin{array}{llllllll}
\partial_{t} u - (\partial_{xx}^2+\partial_{yy}^2) u = 0\\
\mbox{Dirichlet BC}\\
\end{array}
\right. \]
with analytical solution
\[u(t,x,y)=e^{x+y+2t}\]\label{eq:Heat3D1}}

\item{Heat equation 2: 
\[\left\{ \begin{array}{llllllll}
\partial_{t} u -  (\partial_{xx}^2+\partial_{yy}^2) u = 0\\
\mbox{Dirichlet BC}\\
\end{array}
\right.\]
with analytical solution
\[u(t,x,y)=(1-y)e^{x+t}\]\label{eq:Heat3D2}}

\item{Poisson equation 1:
\[\left\{ \begin{array}{llllllll}
\partial_{tt}^2 u + \partial_{xx}^2 u + \partial_{yy}^2 u + 3\pi^2\sin(\pi t)\sin(\pi x)\sin(\pi y) = 0 \\
u\left|_{\partial\Omega}\right.=0
\end{array}
\right. \]
with analytical solution
\[u(t,x,y)=\sin(\pi t)\sin(\pi x)\sin(\pi y)\]\label{eq:Poisson3D1}}

\item{Poisson equation 2: 
\[\left\{ \begin{array}{llllllll}
\partial_{tt}^2 u + \partial_{xx}^2 u + \partial_{yy}^2 u -6 = 0\\
\mbox{Dirichlet BC}\\
\end{array}
\right.\]
with analytical solution
\[u(t,x,y)= u(t,x,y)=t^2+x^2+y^2\]\label{eq:Poisson3D2}}

\item{Poisson equation 3:
\[\left\{ \begin{array}{llllllll}
\partial_{tt}^2 u + \partial_{xx}^2 u + \partial_{yy}^2 u -6 = 0\\
\mbox{Dirichlet BC}\\
\end{array}
\right.\]
with analytical solution
\[u(t,x,y)=t^2+x^2-y^2\]\label{eq:Poisson3D3}} 

\item{Taylor-Green vortex: 
\[\left\{ \begin{array}{llllllll}
\partial_t u + u \partial_x u+ v\partial_y u+\frac{1}{2} e^{-4 t}\sin{2x}-(\partial_{xx}u+\partial_{yy})u=0\\
\partial_t v + u \partial_x v+ v\partial_y v+\frac{1}{2} e^{-4 t}\sin{2y}-(\partial_{xx}v+\partial_{yy}v)=0\\
\partial_x u+ \partial_y v=0\\
\mbox{Dirichlet BC}\\
\end{array}
\right.\]
with analytical solution
\[\left\{
\begin{array}{l}
 u(t,x,y)=\cos(x)\sin(y) e^{-2t}\\
 v(t,x,y)=\sin(x)\cos(y) e^{-2t} \\
\end{array}\right.\]\label{eq:TaylorGreen}}

\item{Lamb-Oseen vortex: 
\[\left\{ \begin{array}{llllllll}
\omega=\partial_x v - \partial_y u\\
\partial_t \omega + u \partial_x \omega+ v\partial_y \omega-5\cdot10^{-3}(\partial_{xx}\omega+\partial_{yy})\omega=0\\
\partial_x u+ \partial_y v=0\\
\end{array}
\right.\]
where 
\[\left\{
\begin{array}{lllll}
\displaystyle w(0,x,y)=\frac{1}{4\pi t}\exp\left[-\frac{x^2+y^2}{4 t}\right]\\
\mbox{Dirichlet BC}\\
\end{array}
\right.\]
with analytical solution
\[\left\{
\begin{array}{l}
\displaystyle u(t,x,y)=-\frac{y}{2\pi(x^2+y^2)}\left(1-\exp\left[-\frac{x^2+y^2}{4 t}\right]\right)\\
\displaystyle v(t,x,y)=\frac{x}{2\pi(x^2+y^2)}\left(1-\exp\left[-\frac{x^2+y^2}{4 t}\right]\right) \\
\end{array}\right.\]\label{eq:LambOseen}}

\item{Vorticity equation: 
\[\left\{ \begin{array}{llllllll}
\omega=\partial_x v - \partial_y u\\
\partial_t \omega + u \partial_x \omega+ v\partial_y \omega-5\cdot10^{-3}(\partial_{xx}\omega+\partial_{yy})\omega-0.75 \left[\sin\left(2\pi(x+y)\right)+\cos\left(2\pi(x+y)\right)\right]=0\\
\partial_x u+ \partial_y v=0\\
\end{array}
\right.\]
where 
\[\left\{
\begin{array}{lllll}
w(0,x,y)=\pi\left[\cos(3\pi x)-\cos(3\pi y)\right]\\
u(t,0,y)=u(t,1,y)\\
u(t,x,0)=u(t,x,1)\\
u(t,0,y)=u(t,1,y)\\
v(t,x,0)=u(t,x,1)\\
\end{array}
\right.\]\label{eq:Vorticity}}

\end{enumerate}

\end{document}